\documentclass[12pt]{amsart}
\usepackage{graphicx}
\usepackage{epsfig}
\usepackage{fancyhdr}
\usepackage{framed}
\usepackage{floatrow}
\usepackage{mdframed}

\usepackage{amsmath,amsfonts,amssymb,amsthm,mathabx}
\theoremstyle{plain}

\newtheorem*{thm*}{Theorem}

\newtheorem*{lem*}{Lemma}

\newtheorem*{cor*}{Corollary}

\newtheorem*{prop*}{Proposition}

\theoremstyle{definition}

\def\\{{\mathcal{R}}}

\def\be{\begin{equation}}
\def\ee{\end{equation}}
\def\ben{\begin{equation*}}
\def\een{\end{equation*}}

\begin{document}

\title{The languages of actions, formal grammars and qualitive modeling of companies}
 %\centerline{\bf  (The product potential social system )}
\author{Vladislav B. Kovchegov}
\email{vlad\_kovchegov@yahoo.com}
\begin{abstract}  

 In this paper we discuss methods of using the language of actions, formal languages, and grammars for qualitative 
conceptual linguistic modeling of companies as technological and human institutions. The main problem following the 
discussion is the problem to find and describe a language structure for external and internal flow of information of 
companies. We anticipate that the language structure of external and internal base flows determine the structure
 of companies. In the structure modeling of an abstract industrial company an internal base flow of information is 
constructed as certain flow of words composed on the theoretical “parts-processes-actions” language.  “The language 
of procedures” is found for an external base flow of information for an insurance company. The formal stochastic 
grammar for the language of procedures is found by statistical methods and is used in understanding the tendencies 
of the health care industry.We present the model of human communications as a random walk on the “semantic tree”. 

\end{abstract}

\date{}
\maketitle


 \centerline{ \bf Introduction }
\par This paper is dedicated to the modeling of an organization. First of all, we would like to clarify what modeling means.  
If a person or groups of people use some process or device, he/she or they may know what kind of “language” this 
process uses. For instance, a lot of people use cars every day. They have to keep in mind a lot of information about
cars; the kinds of sounds and/or vibrations that are normal for the engine and so on. Very often cars have personal 
“characteristics.”  For example, before you start the engine on you may have to pull some wire to get the vehicle 
started. All of these attributes could be considered as a “language” by which the car speaks to you.  More often than 
not it is likely that a driver does not know how a car transfers the chemical energy from the fuel into the mechanical 
movement of the vehicle, but they definitely know something about the cars’ “language.”
\par So, every process or device communicates with us and uses some “language.”  Sometimes this language is very 
primitive, but other times the process speaks to us by using a very complex language.  In this paper we will explore 
the process of interaction between patients and doctors, which is characterized by its own language. This type of 
language (the language of procedures) has a grammar and we will use this formal grammar to generate predictions. 
Simultaneously, this language can be used as a model for professional activities of a company’s employees.  This 
means that in the future we can use self-learning grammar as a model of professional activity.  In the general case 
the processes do not communicate with us directly.  There exist certain technologies and/or devices between a process 
or groups of processes, and a person or groups of people. For example, a car is a device, which transfers chemical energy 
into mechanical movement.  So when we say that “a process communicates with us” this means that a device (a car), which 
uses a given natural process of converting chemical energy into motion communicates with us. In this paper we model 
human illnesses as communication with us through the language of procedures. Thus for us an insurance company is a 
device, which transfers real processes (illnesses) into a language of procedures. For different industries, human beings 
use different technologies and/or devices. So every device is a function of the set of processes and technologies. 
The paper (Kovchegov, 1977) contains information about the “language of technologies.” 
\par The idea to use the formal grammar for strings of action (Chomsky, 1963; Miller and Chomsky, 1963; Novakowska, 
1973a) was developed in the work of Novakowska, (1973a, b), Paun (1976), Scortaru (1977), and Skvoretz,  Fararo (1980). 
Skvoretz (1984) assert that for human action, grammar is either an automaton or a regular type.  This means that for all
sequences of human action there is a finite automaton, which generates the same set of words that was generated by 
the formal grammar. The alphabet, which Skvoretz and  Fararo use, includes the finite numbers of acts and tests.  Among
human actions, actions are not only very important, but all conditions (tests) as well.  As every person knows, human 
actions and the order of human actions depend on a huge number of conditions. This is the main reason why the author 
uses Ianov’s schemes (Ianov, 1958). The second reason is the existence of a canonical form for Ianov’s schemes.
 Ianov’s schemes are invited by Lyapunov, 1958 and Ianov, 1958 as a model of a computer program. Ianov describes 
an algorithm that can recognize when two programs are identical.  Ianov’s schemes can be realized (Rutledge, 1964) as 
a type of automata.   However, not every formal language can be realized as an automaton.
\par The natural languages analyzed bring out many interesting ideas. One of these is the conceptual dependency 
theory (Schank, 1975). The central focus of Schank's theory has been the structure of knowledge, especially in the
context of language understanding. Schank (1975) outlined the contextual dependency theory that deals with the 
representation of meaning in sentences. Building upon this framework, Schank and Abelson (1977) introduced the 
concepts of scripts, plans and themes to handle story-level understanding. Later work (e.g., Schank, 1982,1986) 
elaborates the theory to encompass other aspects of cognition. 
The key element of the conceptual dependency theory is the idea that all conceptualizations can be represented in terms
 of a small number of primitive acts (only eleven!) performed by an actor on an object. The set of primitive acts is divided
 into four groups. The first group is the set of acts performed by human being: PROPEL, MOVE, INGEST, EXPEL (“expel 
something from human being”), GRASP (“grasp an object”). The second group of acts includes two acts: “change position
 of object” (PTRANS) and “change abstract connections between objects” (ATRANSE). The third group consists of SPEAK, 
and ATTEND.   The last group contains two mental acts MTRANS (“mental transfer of information”), MBULD (“create and 
combine thoughts”). Any object has a set of states that are represented by scales. There are a few scales: health, fear, 
anger, mental state, physical state, consciousness, hunger, disgust, and surprise.  Acts must change the state of an object. 
The conceptual cases, rules, categories, conceptual syntax and so on are elaborated as well. The books [6-8] contain not 
only whole descriptions of how conceptual dependency theory works, but more as well.
\par Obviously for the modeling of organizations eleven personal human acts and human feeling scales are not enough. 
We have to increase the list of acts and actions, and add additional new concepts. But Schank’s language of acts and more 
general “language of actions” enable us to make qualitative (not quantitative) approaches to modeling of human activities 
and organizations. 
We plan to use the language of acts and actions (joint or group acts) as a base for the modeling of some features of real 
organizations. So we do not pretend to present a model of whole organizations or some formal model of whole organizations. 
We want to only find the type of language input flow presents and what kind of connections exists between input flow and 
an organization. We try to find a formal description of the language hidden in input flow. It is clear to us that knowledge 
about input flow is not enough for the modeling of a whole company.  If we want to get a whole model of the organization 
we cannot escape from using tools of Informational Technology (we plan all – external and internal- informational flows, 
calculate density, entropy and another informational flows’ characteristics).  So essentially we have a more modest aim. 
The main idea of the presented manuscript is to learn non-traditional languages as a language of meaning (language of actions,
 semantic tree), actions, “parts and actions”, body-gesture language and so on using the traditional formal languages.  
\par Let us express hope that when all languages hidden in external and internal flows of organizations are found we 
may start to do a more exact and formal conceptual model of the whole organization. This, however, is a long-term problem.
\par The second linguistic concept that will be used for modeling human behavior is the semantic tree. Ideally all formal acts
and group actions must change the state of objects and make a trace on the semantic tree or space of a whole company. 
In this article we do only a small step in the desirable direction. The behavioral model of human communication as a random 
walk on the semantic tree is done. In this model we try to use properties of human beings for mathematical modeling and
we understand how weak and controversial this is. The staff, a most hard and controversial object for modeling, is the
 main part of an organization.  
\par In reality we can explain what organizations do by using few words of natural languages.  This does not take very 
long to do. So our main problem is to create the conceptual language that helps us describe what organizations do by 
using some “conceptual code of organization” still not done. Sometimes, however, input language helps us deduce the 
company’s structure (see part-processes-actions language).
\par This article is divided into three parts. The first part is dedicated to the general problems of formalization and modeling 
of a co mpany. Section 1A contains general information about formal languages, automatons, and Ianov’s schemes.  We
then proceed by describing a model of a company. This description can be transformed into mathematical terms. The main 
idea of this article is to represent a model of a company as a flow of words consisting of an unusual alphabet.  For instance, 
using the alphabet of “car parts”, where the car parts consists of car terminology (rims, hood, nuts, etc.).  The letters in 
this alphabet are the name of parts, natural processes, and human actions. The language of shape and the descriptions 
of conditions enable us to make very realistic models of companies.  The connections between conditions for using natural 
processes and shells, that create normal conditions for processes and keep processes in a given range, give the model 
good predictive power. The concept of “semantic space” for industrial activities help us preserve the meaning. The third 
item of section one contains a verbal description of the semantic space and a verbal model of communication. This
description will be used in section two.
\par Section two contains mathematical models of human conversations, and helps us understand the concept of semantic 
space. For a description of human activities we have to obtain information about groups of actions, consisting of actions
that use natural processes/devices that limit them to a given range. These actions give (endue) a person or groups of 
people the ability to control these processes. The last types of actions are actions that strictly use human abilities. Humans 
think, speak, write, read, and so on.  For a formal representation of the type of action, it is helpful to use Ianov’s schemes.  
Formal languages are used to give a formalization of the second type actions.
\par Section three contains a complete description of the procedure’s language and the grammar for input flow of an insurance 
company. The process of interactions between patients and doctors (the main process in this section) communicate with the 
organization that controls it. The main goal of an organization is to keep this process within a given range. So for our device, 
an insurance company is the process that keeps the process within a given range.  To construct this semantic model of input 
flow, we use the alphabet of the types of procedures performed on the patients with the given chronic disease.  For instance, 
the alphabet for diabetes contains 34 “letters.”  The history of the disease might be represented by a short word in a given 
alphabet and appears like “A$\_$ANDR S4 S2”, where A$\_$, AN, $\ldots$ . S4, S2 are letters of the alphabet of the disease.  
The study of information for five years shows that the structure of short words has a tendency to change.  To model this 
tendency we use conditional probability.  The conditional probability is found from the data.  Then using a computer simulation, 
we calculate a set of pseudo-random “short words.”  The next problem is to generate the set of pseudo-random “long words.”  
The long words may appear like “A$\_$-12AN-1DR-17 S4-1 S2-1”, where the number following each “letter” is the frequency of 
encountering this letter.  For this purpose we find the conditional probability Pr {X = “long word” / X = “short word”} and then 
generate a set of pseudo-random “long words.”  The list of “long words” and the list of “normative prices” for procedures give 
us the ability to calculate the mean, “harmonic”, minimal and maximal prices for all diseases.

 \vskip .2in
\par { \bf 1.	THE GENERAL PHILOSOPHY FOR MODELING A COMPANY.  THE LANGUAGE OF ACTION: FORMAL GRAMMAR AND IANOV’S SCHEMES OF A PROGRAM. }
\vskip .1in
\par {\bf  A. TOOLS – FORMAL GRAMMAR, FINITE-STATE MACHINE (AUTOMATA) AND IANOV’S SCHEMES OF PROGRAMS.}
\vskip .05in \rm
\par In this subsection we will describe all the mathematical, or more precisely, computer science or cybernetic tools.  
The formal languages are based on the formal grammar.  The formal grammar G is given by the quadruplet 
$\langle$ N, T, P, S $\rangle$ where S is the root or start symbol, T is the set of terminals symbols, N is the set of non-terminals
symbols, and P is the set of grammar rules, substitutions, or productions.  The alphabet V is the union of T and N, 
where in the general case the alphabet is a set of arbitrary symbols.  The root symbol belongs to the alphabet V.  
Let us use the symbol$V^*$ for all words generated by the alphabet V.  For instance, if 
V=$\{$ a, b $\}$ , then $V^*$=$\{$ Em, a, b, aa, ab, bb, ba, aab, $\ldots$ $\}$ where the empty word set is 
denoted by Em.  The non-terminal symbols correspond to the variables, and the terminal symbols correspond to 
the words of natural language.  The set of grammar rules is the set of expressions a $\rightarrow$ b where a 
and b are words from $V^*$ and the word, a, is not empty.  The formal language generated by the grammar
G is denoted by the symbol L(G).  For example: Suppose we have the grammar G = $\langle$ N, T, P, S$\rangle$, where
 N =$\{$ S $\}$, T =$\{$ a, b $\}$, and P =$\{$ S $\rightarrow$  aSb, S $\rightarrow$ ab $\}$.  We first use 
the substitution a few times to get   S $\rightarrow$ aSb $\rightarrow$ aaSbb  $\rightarrow$
 aaaSbbb $\rightarrow$ aaaaSbbbb $\rightarrow$ $\ldots$.  Using the second rule, we get the set of words generated 
by the formal grammar G (grammar inference): ab, aabb, aaabbb, aaaaabbbb, $\ldots$. 
So, L(G) = $\{$ ab, aabb, aaabbb, aaaabbbb,$\ldots$ $\}$.
\par The grammar rules P defines the types of grammars.  There are a few types of grammars: unrestricted,
 context-free, regular or automata and so on.  The regular (automata) grammar consists of two types of sets 
of productions: A $\rightarrow$ aB and A$\rightarrow$  a, where A, B are non-terminal symbols from N, and 
a is a terminal symbol from T.  A finite-state automata (machine) can realize the regular grammar.
\par The stochastic language is a language generated by stochastic grammar.  The stochastic grammar is the
quintuple $\langle$ N, T, P, Q, S $\rangle$, where N, T, P, S are the set of non-terminal symbols, terminal symbols, product 
and start symbols, respectively, and Q is the set of probabilities on P.  So, if word x can be generated from 
the start symbol by the set of grammar rules $r_1$, $\ldots$,$ r_N$, then the probability of word x to be
generated is p(x) = p($r_1$) p($r_2$$\vert$$r_1$) p($r_3$$\vert$$r_1r_2$) $\ldots$ p($r_N$$\vert$$r_1r_2 $\ldots$ r_{N-1}$), 
where p($r_K$$\vert$$r_1 r_2\ldots  r_{K-1}$) is the conditional probability to use the product $r_K$  if the rules 
$r_1$,$ r_2$,$\ldots$,$ r_{K-1}$ are used before.  In this article we will use stochastic grammar for
modeling insurance company input flow.
\par Example how linguists use stochastic grammars. Data Oriented Parsing (DOP) models for natural 
languages analysis provides very interesting example of application stochastic grammar. The DOP 
model was first introduced by Remko Scha (Scha, 1990) and formalized by Bod (Bod, 1992). DOP 
method has used training set of statements (“training corpus”) for generation the special stochastic 
grammar. Then delivered stochastic grammar will be used for providing semantic analysis. Right now 
we describe how stochastic grammar will be generated from training corpus by one version of DOP 
model (Rena Bod, Remko Scha and Khalil Sima’an “Introductions to Data-Oriented Parsing”  2001). 
Every statement from training corpus is source for “parse tree”, where every node can be terminal 
or non-terminal. Every terminal node is labeled by lexical type (there are lexical types for the verbs, 
for noun and so on). Every non-terminal node will be labeled by “head-complement rules” (“root’s labels”) 
that necessarily for future decomposition. Set all lexical types is terminal set T. Set all “root’s labels” is set
 of non-terminal symbols N.  Then all parse trees will be decomposed into set “elementary trees of flexible 
size” (product set P) and will be found frequencies of all elementary trees with common root label in whole 
populations (the set of probabilities on P). So we have set of terminal symbol T, set non-terminal symbol N,
product set P and the set of probabilities on elementary trees Q. When we add start symbol S we get whole 
stochastic grammar. Very likely that for two different training set of statements we will get two not completely 
identical “alphabets” (where “letters” are elementary trees)! 
\par The deterministic finite-state machine is the quintuple $\langle$ S, I, O, f, h$\rangle$, where S, I, and O represent 
a set of states, a set of inputs (input alphabet), and a set of outputs (output alphabet), respectively, 
and f and h represent the next-state and output functions.  The next-state function, f, is defined 
abstractly as mapping the cross products of S and I into S.  In other words, f is assigned to every 
pair of state and the input letter is assigned another state symbol.  So, the next-state function, f, 
defines what state (of the finite-state machine) will be in the next time interval given the state and 
the input values in the present time interval.The output function h determines the output values in the 
present state.  There are two different types finite-state machines, which correspond to two different 
definitions of the output functions h.  One type is the Moore machine, for which h assigns an output
symbol to each state of machine. The other type is the Mealy machine, for which h is defined as a 
function that assigns every pair of state and input letter an output letter.
\par There are many different types of automata (deterministic and non-deterministic finite-state machine) 
and automata networks (for instance neuron-like networks) that may be used for modeling, but for now 
we have to define Ianov’s scheme. Lyapunov and Ianov’s model of a computer program is what invites 
Ianov’s schemes.  Ianov describes an algorithm that can recognize when two programs are identical (Ianov, 1964).  
This job uses very abstract language and we can try to use this for human action as well.  Ianov’s schemes of a 
program can be defined as a connected oriented graph G, where every node is an operator (action) or a 
special device that recognizes either conditions (we call it a Recognizer) or only the Stop Operator. Every
Operator transfers all its memory onto all memory.  Every Recognizer is a logical statement defined by 
the set of parameters.  These parameters need to be described. The canonical or matrix form of 
Ianov’s scheme with n different operators A(1), $\ldots$ , A(n) is Ianov’s scheme which has (n + 1)(n +1)
Recognizers R(i,j) (indexes i and j take values from set 1,2, $\ldots$ ,n+1) and every Recognizer has one input 
and two outputs: one arrow goes to the next recognizer (from R(i,j) to R(i,j+1); let R(i,n+1) equal true for all 
values) and the second one goes to the Operator A(i). If the condition is true, then we must use an arrow from 
the Recognizer to the Operator (all recognizers R(i,n) have arrows to the Stop operator).  Otherwise we have 
to use arrows that go from Recognizer to Recognizer.  The recognizer R(0,j) is the input recognizer, and R(0,j) 
(for all j) has the input arrow from operator A(j). The condition, F(i,j), for the Recognizer, R(i,j), has to satisfy 
the next condition: F(i,j)$\&$ F(i,k)= true, if j does not equal k . Rutlledge, 1964 has defined the canonical form 
for the program and proved that the two programs do the same if and only if they have the same canonical forms.
 He also found the finite non-deterministic automata that do exactly what Ianov’s schemes do.  It is a very
 attractive theorem. This theorem gives us a good ability for recognizing if the two systems of action differ.  
It is very constructive way, but this theorem is not true for all type of operators. The operators have to be 
independent and map the same space onto the same space.
\par Anyone who wants to apply this formal language, either automaton or Ianov’s schemes, to social 
problem meet at least a few obstacles.  The first problem is how to define all actions as an operator on
a few sets (an operator maps one set to another).  For instance, the algebraic group of the isometric 
operator is a very good tool for modeling all kinds of movement.  The question is: how can we describe 
a moving action with damage?  The second problem is how to describe who did actions and what was 
the object of action.  Sometimes in order to make an action we have to use a lot of people simultaneously.  
The number of people and/or devices that have to participate in the actions simultaneously is a very
important characteristic of the action.  If the number equals n, then we state that the action is n-tuples actions.
\par We will now describe a general model of organization based on the parts – processes – actions language.
\vskip .2in
\par { \bf B. THE BASE MODEL OF AN INDUSTRIAL COMPANY }
\vskip .1in
\par For simplicity we divide all companies based on the industrial and informational type.  
In real life all companies have industrial and informational parts.  The industrial part consists of all 
activities under material flow, while the informational part has business with information. 
However, informational parts also include material flow such as: papers, envelopes, phones, 
computers (PC and mainframe), communication’s networks and so on.  We can then divide all 
activities on the internal and external parts. The internal parts focus on internal technological 
flows in the company and the informational system has to reflect the current condition of industrial flows.  
This information will be used for the operational decision making process. The strategic and tactical 
decision is to make processes need more external information and create a model of the external word.  
In this article we concentrate only on the internal part of a company’s activities.
\par For the modeling of internal industrial flows we separate two types of actions: assemble and 
disassemble.  We can find traces of this type of action almost everywhere.  For the formulization 
of this type of action we can use a formal language as well.  If a person or groups of people want 
to assemble something, he/she or they need elements that will be use as parts.  Some of these 
parts are assembled somewhere and are just labeled as parts.  While other parts are done in a 
company and may be the result of some activities.  For these parts we will use the alphabet of 
actions using special symbols for denoting tools and natural processes – devices.  Any device can 
represent a main natural process, so when we say devise we mean to keep in mind some natural process and vice versa.
\par So, let us denote E, the set of elements: E =$\{$ $x_1$, $\ldots$ , $x_N$ $\}$, where N is the 
number of parts. For instance, for a car the number of parts N equals approximately fifty thousand.  
We then describe the set of methods for gluing the elements and the set of human actions that do this.  
Let us use the symbol P for the set of gluing processes and/or devices and A for the set of actions of gluing.  
So, if the process is mechanical, we need parts and facilities (wrenches, bolts and so on).  If the process is 
chemical we then have to describe the type of glues, conditions and devices.  The whole assemble-action 
(A-action) appears as the word 
(($x_1$,$x_2$: $a_1$,$a_2$, $P_1$), $x_2$, $x_4$, ($x_5$, $x_4$: $a_3$, $a_5$, $p_2$, $p_5$); $a_4$, $a_3$, $a_7$, $p_3$), 
where the elements of E: the action from the set of action A,$p_1$, $\ldots$, $p_5$ belong to the set of 
processes P (chemical, physical, biological and so on). The parenthesis denotes we get a new element.  
For instance, the word  ($x_1$, $x_2$: $a_1$, $a_2$, $p_1$) represents the new element assembled 
from $x_1$, $x_2$; and for gluing uses process $p_1$ from P and human actions $a_1$ and $a_2$.  
But all elements must satisfy the same conditions. We can describe the set of conditions and denote 
this set by the symbol C.  So, we say that elements xk belongs to C($x_k$), if $x_k$ satisfies the set 
of conditions C($x_k$).  Similarly, ($x_1$,$x_2$:$a_1$,$a_2$, $p_1$) must belong to C($x_1$,$x_2$:$a_1$,$a_2$,$p_1$) and so on.
\par Disassemble actions (D-actions) transform one element into the set of elements or fractions. 
For instance, D(x) =$\{ Fx_1, \ldots , Fx_n \}$, where x is the initial object and $Fx_1, \ldots , Fx_n$ 
are fractions of x.  Symbol F is the first letter of the word “fraction.”  Occasionally it is better to use 
the distribution of debris sizes. Sometimes, the result of D-actions is the set of parts or parts and fractions. 
 D-actions can be realized by using different natural and artificial processes: from explosive materials to
mechanical processes.  So, D-actions can create fragments, debris and parts.  Results of D-actions can 
be used for assembling processes. If object x consists of a biological nature, the result of D-action is 
a type of injury.
\par Thus, every element (“letter”) has a shape and/or weight, material and as a description, these 
things (shape and so on) need to use a special language as well.  For example: a description of the 
shape is a cylinder with diameter 2.456 feet, height 3.2 feet, depth of wall 1.2 inches; weight 
1.03 ton; steel.  If the element is a bit of information, we use another description: 12 millions records, 
length of record is 238 symbols, MS Word (extension .doc).  But for our models, this information must 
be extended by the place of information on the semantic space (see below).  If the element consists 
of information we need to explain what the information is about.  For the modeling of a company we
 build the semantic space (tree) and all processes and actions must have the informational presentation 
on the semantic space. The creation of a semantic space for particular case is a very big problem.  
The high level of a semantic tree for a large number of tools can be expressed by the scheme  
“engine – transmitter – working tool - control.”  The typical engine transforms chemical energy into mechanical, 
the transmitter propagates mechanical action to the working tool, and the control system helps to control 
the working process.  This scheme, however, is not unique and these set of schemes generate a 
technological semantic space.  We will use a semantic space as a system of coordinate for all human products.
\par Thus following, for any process (mechanical, physical, chemical, biological, social and so on) 
we have to describe the condition and distance from the “normal” condition.  We describe the normal
 condition as conditions normal for humans: physical, chemical, biological conditions (consistency of air,
gravity, radiation, temperature and so on), landscape and green/animal words as well as demographic, 
social, and other conditions.  Some conditions are considered good for processes but wrong for humans.  
In order to use these processes people create a shell, known as the shell philosophy.  For deeper modeling of
 all processes we need to use the shell language.  Examples of popular shells are homes, and clothing.  
These types of shells protect people from bad weather and decorate them.  Examples of industrial shells 
are chemical and nuclear reactors.  An ordinary vehicle consists of a combination of a few shells: the 
cabin of car protects the driver and the passengers. The engine creates the condition for burning 
fuel and transforms the chemical energy into mechanical.  While a bathyscaphe protects people 
from high pressure and so on. Sometimes it is very easy to predict the shape of the shell and/or 
materials used.  For instance, all devices that protect people from high pressure contain a firm 
steel camera. We can find similar examples. Friendship may be interpreted as a human wish to create
a social shell.  It is easy to view people as being members of a group by their participation of an action,
 their ability to create the “tools” for the action, and by the set of  “shells” that support the normal condition 
for the person, the group, and the whole society. We cannot say, however, by definition that all people 
are members of a team that build a reactor for a chemical process.  People, however, try to surround 
themselves by friendly people and attempt to create a social shell.  The purpose of this is to create a 
normal condition for the one person or the whole group.  The social aspects of a company will be discussed in another paper.
\par For now we can rewrite the semantic scheme for tools that move and carry (cars, airplanes, ships
 and so on) by the description “engine – transmitter – working tool – control - shells.”  Chemical processes 
(including some metallurgical processes) use high-level semantic schemes such as “load to pot –
chemical ‘cooking’ – unload pot,” where “pot” is the shell for reactions.  How can we describe a shell?
We can think of a shell characterized by condition, shape and materials.  Simultaneously, we must 
find a place of the shell on the semantic space.
\vskip .2in
\par { \bf DESCRIPTIONS OF THE BASE OF A COMPANY }
\vskip .1in
\par The assembly word gives us information about the structure point of assembly, facilities and communications.  
We can define the physical body of a company.  Parentheses represent not only objects, but points of 
assembly as well. There are many types of assembly locations varying from primitive to very sophisticate 
devices.  Following this, the systems of parentheses give us information about order of actions. If a
 parenthesis is within another parenthesis, this object has to be done first before the second.  If we 
have information about the assembling time for all objects, we can figure out the structure of jobs and 
can estimate the number simultaneously working spots.
\par Suppose, we have A-word 
$$((x_1,x_2;a_1,a_2, P_1), x_2, x_4, (x_5,x_4; a_3,a_5,p_2,p_5); a_4, a_3, a_7, p_3)$$ 
and suppose the assembling time for object one $(x_1,x_2;a_1,a_2, P_1)$ is $t_1$, for object two 
$(x_5,x_4; a_3,a_5,p_2,p_5)$ the assembling time is $t_2$, and for the whole object the assembling time is 
$t_3$, where $t_2$ approximately equals 2$t_3$, $t_1$=3$t_3$.  In this case for a continuous job we need 
3 assembly locations for object one and two assembly locations for object two.  If we know the number of
workers working on object one we can estimate the necessary number of employees needed to complete the task. 
We then have to calculate the number of managers needed and we can finish the evaluation of necessary 
employees for all A-, D-operations.
\par All of the information reflected by the current situation of A and D actions need to be gathered by managers as well. 
The informational part of the model of a company will be discussed later when we describe the almost pure 
informational company – insurance company. If we change proportions between the assembling times 
($t_1$=x $t_3$, $t_2$=y $t_3$, where x and y not integers) we can get a more complex situation and 
find the necessary operative control.  In the general case x and y are random numbers and these objects 
cannot satisfy the necessary condition.  As a result, our managers are given the jobs of operative control.
For a formal modeling of the operational control we can use Ianov’s schemes.  In the general case it is not 
a small job to prepare the scheme.  We need to describe all possible conditions and manager reactions.  
It is possible, however, for either an A - or D - action to do this and then combine these schemes into 
one big scheme and optimize it.
\par The shape, weight, and materials of objects in words determine which objects stay still and which objects move. 
The cumbersome and/or fragile and/or heavy objects are more likely stay still than be relocated for the 
next step of the assembly process.  The maximum sizes of the objects are defined by the size of the building 
(the main shell of a company), and the connections between the objects define the structure of the 
rooms and corridors.
\par For us the model of an industrial company or an industrial base of a company is the flow 
of A- and D – words in parts – processes –actions alphabets saturated by information about 
the shape and conditions.  From this flow we can acquire a lot of information about the company’s 
structure, the necessary connections between the points of assembly/disassembly and the number 
employees.  This flow then gives us information about architectural features of the building (the main “shell” of a company).
\par We can then proceed onto the next step in the modeling of company.  We will describe the 
semantic space as a base for pure human actions as conversations, conflicts, and so on. For 
this purpose we will describe the model of restaurant.
\vskip .2in
\par { \bf  C. HUMAN ACTIONS AND SEMANTIC SPACE}
\vskip .1in 
\par For operations restricted under certain information, we need to describe the semantic 
space (field) of the action. For example, if we want to describe a person’s responsibilities 
we need to create a list of actions and conditions (often times responsibilities depend on 
the situation).  These descriptions have two parts: official and not official.  Given a restaurant, 
a waiter’s responsibilities officially appear to be very easy.  A waiter needs to (1) prepare 
a place for clientele, (2) take orders from clientele, (3) pass on requested orders to the 
kitchen, (4) serve dishes to clientele, and (5) take money from clientele. Unofficially, the 
role of a waiter consists of other things. For instance, a waiter needs to be polite with clients. 
How can we describe the behavior of a person, particularly a polite behavior?  There are 
similar problems for all social actions.  Suppose we want to describe the act of talking. We 
first need to describe a semantic space (spaces).  This arises the following questions. What
 do we currently mean by semantic space?  How can we describe semantic spaces? The semantic 
space can be professional and nonprofessional.  Professional talk can often be interrupted by 
nonprofessional themes (the current news, family problems, sports, celebrities, internal 
relationships, and so on). For a rough description of the semantic space we can only indicate 
the name of the theme or themes of talk (a level one).  We then have to extend our description. 
Descriptions can contain many levels. Sometimes not only are the content of information (oral or 
writing) important, but the type of form it has as well.  For instance, in the case of conflicting 
talk, letters, and so on we need to be extremely accurate.  In this paper we model a type of 
conflict: if a word (the word of procedure’s language) does not belong to a normative set 
something is wrong.  In the case of conflict, the parties (regularly two parties) can appeal 
to the rules, agreements, state and/or federal and/or international laws, norms of morality, 
and so on.  But regardless, in the case of conflict there is appeal.  In our case of a “waiter – client”
 the normal situation may easily be transformed into a conflicting situation. The sours of conflict 
are a very important element of the semantic space of conflict.  It is not easy but possible to 
make a universal description of the semantic space of conflict.  In the case of the “waiter – client” 
the sours of conflict may involve the bad job performed by the crew, the mood of the clients, 
and so on. At any rate, if we want to make a working model of the social action we need to 
providently add the conflict regime to our description.  For instance, we must keep in mind 
that a waiter can call the police or a drunken client can hit someone. We need to extend 
our description in this direction as well.  Our description needs to contain information about 
the number of actors in an action, some information about all of the actors, the kinds of 
actions performed by the actors (“Mr. X hits Mr. Y” or  “the client actions contains a threat”), 
and their physical state (“a drunken client”, “a young boy”).  Other useful information would 
be the personal style of actors (“a rough waiter”, “a gentle person”, “very professional person”).
\par The collective or personal action maps the semantic space onto itself.  This means that 
we can use the canonical form of Ianov’s scheme of action. The action has to change the 
state of objects and/or the state of the semantic space.  The meeting (collective action) 
has to change the semantic space of the theme and change the person’s states (they may 
obtain additional information about a discussed object while simultaneously feel tired).

 \vskip .2in
\par { \bf  i)	THE MASK-POSE-GESTICULATION LANGUAGE.}
\vskip .1in

\par Humans use an assortment of different languages (see Darwin, 1872, Fast, 1970). 
 Firstly they use natural oral languages, followed by professional languages (for instance, 
mathematics is labeled in the set of special languages), physical languages such as sign 
language for the speech impaired, and so on.  We will concentrate our attention on the 
mask-pose-gesticulation language (MPG-language).  There are many types of masks.  
A person can facially express scales of sense varying from seriousness to liveliness.  Every
one knows and uses this “mask” language and it is considered to be a rich language.  
Often people (first of all an emotional person) cannot control their facial expressions and 
everyone can read the language of their mask.  This means that the reaction of the person 
on the word or event can be read.  So the mask-language is very informative and plays a 
very important role in human society.  Similarly, the “language of the pose” may be used 
for different aims. For instance, a person can express respect or contempt, independence 
or dependence, and so on thru their pose.  We can easily recognize hand movement as well. 
All three are composed of some alphabet and, perhaps, some form of grammar.  What is 
important for us is the combination of the elements of the sign language.  For instance, the
pose expressing submission that contradicts with an insolent mask on the face.  So, if we have 
the a of masks M=$\{$ m(1),$\ldots$ , m(N) $\}$, a set of poses P=$\{$ p(1),$\ldots$ ,p(K)$\}$, a set of  
gesticulations (hand movements)  H = $\{$ h(1), $\ldots$ , h(R) $\}$, and a set of themes 
T= $\{$ t(1), $\ldots$ , t(S) $\}$, we can then describe the grammar of mask-pose-gesticulation language 
(MPG-language) as a subset Gr of the (Cartesian) production T$\times$M$\times$P$\times$H.  The themes set 
represent the semantic space (field) of the situation or action.  This means that if we can 
describe the situation by these sets t(i), there is then the set of admissible combination 
(t(i), m1, p1, h1), $\ldots$ ,(p(i), mL, pL, hL) from the set Gr.  The grammar Gr depends on culture
 and other factors. The personal style is the distribution of probability on 
(t(i), m1, p1, h1), $\ldots$ ,(p(i), mL, pL, hL) for a given t(i).  For a given situation, if a person can
demonstrate an unusual behavior this means that the behavior combination does not belong to Gr.
We can generalize the MPG-language.  If we add the action language we obtain the Behavior language.
\vskip .2in
\par { \bf  ii)	HUMAN ACTIONS AND FUNCTIONS.}
\vskip .1in
\par We finish this part of paper by presenting an example of how we can use 
formal language to model human actions and functions.  We cannot do this in a common case, 
but we will at least describe the set of problems.  Every person can do a limited number of 
physical actions and substantially more mental actions.  Mankind lives in an artificial world.  
We use many different tools and networks of tools: cars and road networks, telephones 
and telephone networks, computers and computer networks, the set of network that 
distribute and carry different kinds of goods and so on.  We can add to this list other
networks such as power networks (for instance, network of precincts), healthcare networks 
(the set of networks of doctor offices), the networks of department stores, and so on. 
 The combination of human actions and tools give us unusual results.  We thus have to 
define an alphabet of primitive human actions (physical and mental). What is meant by a 
“primitive action?”  This depends on the problem. We have to find a good descriptive 
level for a given problem.  Before a person starts an action they have to recognize 
distinct situations.  How can we divide this process on elementary actions?  What kind of 
symbol or sign should we use for the identification of social situations?  Humans can recognize 
social situations, but how can they describe “the social landscape” of a given society?  
How can we transform the local situation into the relatively global situation?
\par An example. This example is borrowed from articles and Schank, 1977 and Skvoretz,
J.  Fararo, T. J., 1980. Suppose we have similar objects: a restaurant, waiters, a kitchen, 
and a flow of clients.  In this case we have to describe the waiters functions. First of all the 
waiter needs to keep in mind a lot of information about the clientele. The waiter has to 
recognize what kinds of people arrived to the restaurant. The farthest behavior of the 
waiter is dependent on the result of this recognition.  As a result the number of the 
waiter’s actions is not so big.  He has to help in finding a comfortable place (P), serve 
the table (J), setup the chairs (H), bring the menu (B), take the order (T), pass the order 
onto the kitchen (K), bring appetizers (A), bring dishes (D), take money (C), wait (W), 
call the manager, call the power network (N), say good-bye (G). The control and recognition 
actions are: make the recognition (R), dress and wear a mask (M), discuss the contents of the 
menu and help patrons make right choices (S), check the result of the action (O), trace customers 
constantly (Z).  It is easy to see that the list of “elementary actions” may contain very 
complicated actions.  For instance, the action R (recognition) is not an example of an 
“elementary action.”  Action Z (to constantly trace customers) is not action.  
It is a sub-function.  If a customer (group of customers) is (are) drunk, aggressive, 
the waiter may then call the manager or police.  In this case we have a conflict situation.  T
here are a lot of other reasons for conflicts such as: bad service or food, a dirty table, 
wrong calculations on the bill, and so on. There may possibly be conflicts between the 
clients, too.  For a description of conflict we need to use another list of actions.  
This means that we have to create the universal model for a conflict situation. We then 
have to appeal to higher level of descriptions.  In this case we have a case where 
the local situation can be transformed into a relatively global situation.  
The regular mask for a waiter is the mask of a “hearty welcome”(M1) or “friendly, 
cordial, open-arm guy/girl” (B2). Let symbol M2 denote a bad mask.  There exists, 
however, a personal style for the waiter and he/she has to arrange their mask for 
a particular type of consumer.  Some clients like short distances, some clients do 
not like familiar relations and so on.  The waiter must recognize (action R) the type 
of client he/she has gotten and the waiter has to correct his/her behavior (make mask).
\par So, the typical fragment of a working day for the waiter can be represented by the following sequence of letters:
\par U1:	M1ZR FOHOBOSOTKOAOZDODODOOCG
\par U2:	M1R FOHOBOSOOTKOAOZDODZODOC
\par		M1RM2”Conflict”
\par U3:	M2R FOHOBBOSOTKOAOZDODODOOCG
where U1, U2 and U3 are service units.  The service unit is a person or a group 
of people that have to be served together. What do we mean by fragments?  
We can easily explain the above sequence.  The waiter gets a unit of service 
(U1), he/she dresses the mask of a “hearty welcome” (B1) and tries to help find 
a good place for the customers to sit (F). The waiter then brings additional chairs 
if needed (H), controls how the client (clients) feels and brings them the menu (B), 
and so on.  We can see a lot of embedded O’s and Z’s. It reflects that O (check result 
of action) and Z (constantly trace customers) are functions. O’s and Z’s are not so 
different: in case O, the waiter has to communicate with the clients personally, in 
case Z the water just looks at the clients.  Our function is a set of actions that 
need to be repeated constantly.  From this point of view the waiter’s complete 
job is just a function.
\par But for the modeling of a company (restaurant for instance) we 
have to describe the input flow.  In the previous example it is a flow of clients. 
 What kinds of parameters are essential for this description? The answer 
depends on the problem.  For the flow of clients the important thing is the 
average time between two incomes.  Following this is the importance of
the average size of the customer’s group. It is then important the state and/or 
behavior of the client.  For a profit organization the importance lies in the 
paying-capacity of the person.  The clientele cannot be served if they do 
not satisfy a certain set of conditions. For a movie customer of an R rated 
movie they have to be older than 18 years of age.  For a drunk, aggressive 
person all companies are practically closed.  So, if every person can be described 
by a vector x=(x1, x2, $\ldots$ , xN), then a restricted condition appears as 
“x belongs to set G”, where G has to be described by the laws and by the 
particular company.  For instance, a drunk and/or aggressive person would be 
a good client for a precinct.
\par This kind of model gives us the ability to get more detailed information 
about different views of company life.  For this purpose we need to 
add more details to our description and information about our main processes. 
In the case of the restaurant we need to add additional information about 
a unit of service. For instance, we have a sequence of global, local, international, 
and so on events: E1(t1), E2(t2),$\ldots$, Em(tm), where t1, t2,$\ldots$ , tm is the 
sequence of time points over the past few days (t1 < t2 <$\ldots$ < tm). Suppose 
that E1 is a football game between two very popular teams and let us consider 
that we have a flow of events (E1, sport), (E2, entertainment), 
(E3, politics + federal), (E4, sport), (E5, interfamily relations) and so on, 
where the first word is the code of events and second represents a type 
of event or a code of the type of evens on the semantic scale.  In this case 
the semantic scale (field) is just a list of items.  It is very likely that the 
contemporary events are an important source for the themes of conversation 
(at least as a starting point for conversation).  For every person there exist 
the flow of the personal events.  Sources of personal events may be health 
(how you feel), family, job, and so on.
\par In the modeling of a conversation we need additional information about 
the person. Everyone has a set of favorite themes, F, and a set of  “sick” 
themes, S.  A theme liked by the person is labeled favorite.  The sick theme is
 a theme that evokes an unusual reaction.  Every person has a personal level 
of activity in a conversation (coefficient of aggressiveness k).  If we have m 
conversation partners, we have to describe sets F and S for all participants. 
 Let us denote by symbols F(i), S(i), and k(i) the favorite set, the sick set, and 
the coefficient of aggressiveness for person i, accordingly. 
 F(i) =$\{$t(i,1), $\ldots$ , t(i,ki) $\}$, S(i) =$\{$ s(i,1),  $\ldots$ , s(i,ni) $\}$,
 where t(i,1), $\ldots$ , t(i,ki), s(i,1), $\ldots$ , s(i,ni) 
are the set of theme names or theme codes on the semantic scale; and ki , ni 
are the number of themes in F and S, accordingly.  The intersection of sets F(i) 
and S(i) cannot be empty.  For some people the set S can be a subset of F and so on. 
 Suppose we have a flow of “breaking news” (P1, pt1), (P2, pt2), $\ldots$ , (PN, ptN), 
where the first word is a description of events, the second one is the name of 
the theme or code of the theme on the semantic scale.  For instance, the hot 
information can look like (“Team A knock out the team B”; “sport, local”). The first 
statement gives us information about the event; the second statement is the 
name of the theme (“sport”) and the level of the event (“local”).  If we have 
a free conversation, where the theme of the conversation is does not determined, 
we must define a support function sup(j,t) for theme t and person j.  We assume 
that the arbitrary theme can be supported or rejected as a subject of conversation
 by participants.  For a normal situation if the theme is considered “sick” by at 
least one person, it will be rejected.  During a free conversation participants 
will normally from time to time skip themes. The second phenomenon associated 
with free conversation is that it constantly appears and disappears from 
conversational subgroups. To formalize both properties of a conversation 
process we will use a support function and the structure of F and S sets of themes. 
\par We define the support function for person j:
\par sup(j,t) equals 1 if  theme t belongs to F(j),
\par sup(j,t) equals -1 if  t belongs to S(j)\ F(j) (t belongs to S(j) and does not belong to F(j)),
\par sup(j,t) equals 0 if  t does not belong to F(j) or S(j).
Let us denote using symbol Sub(t) the set of people for whom a support 
function equals one for theme t. This set is a potential candidate to be a 
conversation subgroup if the person moves freely, but in the same case w
e have to use the neighbor’s function N(j), where N(j) is the set of neighbors of person j.
 The neighbor’s function for twelve people and a rectangular table (two long sides 
with five chairs on either side, plus two chairs for the two small sides) looks like
 N(1)=$\{$1,3,8,4,9 $\}$, N(2)=$\{$2, 7, 12, 6, 11$\}$, N(3) = $\{$3, 1, 4, 8, 9$\}$,
 N(4)=$\{$4, 3, 5, 9, 8, 10$\}$, N(5) = $\{$ 5, 4, 6, 10, 9, 11 $\}$, 
N(6) = $\{$ 6, 5, 7, 11, 10, 12, 2$\}$, N(7) = $\{$7, 6, 2, 12, 11$\}$, 
N(8) = $\{$ 8, 1, 9, 3, 4 $\}$, N(9) = $\{$ 9, 8, 10, 4, 3, 5, 1 $\}$, 
N(10) = $\{$10, 9, 11, 5, 4, 6 $\}$, N(11) = $\{$ 11, 10, 12, 6, 5, 7 $\}$,
 N(12) =$\{$12, 2,11, 7, 6 $\}$. 
 So, all processes of free communication will be divided into twelve parts
and for every particular j there can run N(j).  We can then describe 
the structure of intersection of interests for neighbors (intersection for 
favorites themes).  However we first have to use the set of common 
themes (interesting life stories, anecdotes, rumors, and so on).  
This means that every set of people have common themes for conversations.  
But different social strata use different types of interesting life stories, 
anecdotes, and rumors.  The list of possible situations for six people is shown on the transition graph. 
The transition graph for nonzero coefficients of intense of probabilities to 
jump from one partition to another for small period of time (Pr(X(T=s)=D(k)/X(T)=D(j)) = m(j,k)s + o(s)) 
 is $\{$ ( (1,2), (2,1), (1,3), (3,1), (1,4), (4,1), (1,5), (5,1), (1,6), (6,1), (1,7), (7,1), (8,1), (1,9), 
(1,10), (10,1), (1,11), (11,1), (1,12), (12,1), (1,13), (13,1), (1,14), (14,1), (6,12), (12,6), (7,12), 
(12,7), (8,13), (13,8), (9,13), (13,9), (10,14), (14,10), (11,14), (14,11) $\}$.
 \par So in the case of six partners we have various situations.  Case z means 
that three people have the same set of common themes, and the next 
three pairs have common interests as well.  But they do not have common
 interests for other combinations.  This means that real conversations are
 divided by the next two conversation subgroups.
\par The model of free conversation uses few sources of themes: 
flow of current events (global, local, personal), the property of human 
memory (associative), and themes borrowed from neighbors (if the 
intersection N(j) and N(k) are not empty, the theme from one subgroup 
can borrow another one). If the intersection of sets Sub(t) and N(j) 
for the same theme t and subgroup N contain more than one element,
 it is more likely for people to keep talking about these events, the 
problems inside theme t than to skip it.
The last source of information (for modeling a professional meeting) is 
the natural process as a sequence of events. If an event occurs we also get the after-effects.  
For instance, if the government tries to ban alcohol consumption and issues 
a “dry law”, a drug problem will follow.  The experienced person may explain 
by the given events what kind of after-effects and/or back effects will result.
So, for us the “natural process” is the set of sequences of events or words in 
the events alphabet.  Some sequences are more likely than others.  
From the language point of view the main problem with the description on 
a natural process is to find a good alphabet of events and find the formal 
grammar that generates the sequence of words (the sequence of events) 
with some nonzero probability.  If a conversation partner recognizes which 
process generates a given event, he/she can create the next step of forecasting.
There are sets of popular themes.  When people do not know appropriate 
conversation topics they use the theme of “Weather” or “Travel/Tourism.” 
 For others, popular themes are “Sports”, “Politics”, “Hunting/Fishing”, 
“Shopping”, “Family matters”, “Human relations”, “Love”, “Rumors”, and so on. 
 In a traditional society the first three themes are traditional themes for men, 
while the rest of the list represents traditional women themes.  We use some 
code in the representation of a real theme.  We use a word for the 
representation of a narrow theme, so for this purpose we create an alphabet.  
If letter S represents “Sport” and F represent “Football”, then theme “Football”
 is represented by word SF and so on.  It is important to distinguish the
 following: two people may love sports, but one may only love football while 
the second one loves swimming and hates football.  It follows that the longer 
the word the more details about the theme it contains.
\par An example.  Suppose we have three people and three favorite sets 
F(1) =$\{$a, b, c $\}$, F(2) = $\{$ a, b, d $\}$, F(3) = $\{$ d, c, l, m $\}$, 
where a, b, $\ldots$ ,l, m are the set of words and every word in some alphabet 
represents some theme.  These sets are second level information and can
be used by strangers.  Suppose that every person has the coefficient of 
activity x, where x is a non-negative number.  Vector (x1, x2, x3) is the 
vector of activity for all three people.  In the case of the restaurant we 
assume that a drink can change the value of the coefficient of activity. 
 Every person then arranges the set of favorite themes F.  In the general case 
this arrangement depends on the situation and reflects the persons’ 
understanding of what is more or less suitable for a given situation. 
 The arrangement is the set of numbers w(1,1), w(1,2), w(1,2), where weight w(1,1) 
is larger than w(1,2) if and only if the theme is more suitable than theme b 
for person one.  For person three the arrangement is the set of weights
 w(3,1), w(3,2), w(3,2), w(3,3) because of F(3).  Set F(3) contains four themes.  
The probability that person j proposes a theme for conversation
 equals xj/(x1 + x2 + x3), where j belongs to set $\{$1, 2, 3 $\}$.  The conditional probability 
that theme t is supported by partners must be proportional to the number of
 people supporting theme t minus one.  In our case 
P{a$\vert$1} = (2-1)(w(1,1)/(w(1,1) + w(1,2) + w(1,3)))6, because sub(a)=$\{$1,2$\}$.  
Suppose that for all people the sum of weights equals 1 and all weights are non-negative. 
 In this case the general formula for the calculation of the conditional probability 
for theme t to be supported by participants is:
P$\{$ t$\vert$j $\}$ =v ($\vert$sub(t)$\vert$ - 1)w(j,k)/$\vert$F(j)$\vert$,
where theme t belongs to F(j), contains number k and $\vert$F(j)$\vert$ is the number 
of elements in set F(j).  Symbol $\vert$sub(t)$\vert$ has the same meaning (the number elements).  
Coefficient v must be found from the condition: the sum of all P{t$\vert$j}for
 all themes t from F(j) has to equal one. Then in random time the conversation 
theme may be skipped.  This means that when a theme of conversation is 
exhausted it needs to be changed.
\par For a more realistic model we have to add the set of “sick” themes.  
Suppose that the “sick” set for person one is S(1) = $\{$ c, j, k $\}$, the set of 
 the “sick” problems for person two is S(2) = $\{$h $\}$, and S(3) = $\{$ c $\}$ 
is set of sick problems for person three.  Then suppose that every person has 
an opinion about a “sick” theme. In this model we will use two polar opinions.  
The first one is “I completely support and like”, the second one is the opposite 
“I completely do not support and hate.”  If person one or three proposes theme 
c as the conversation theme and if they differ in opinion about similar events
 (any theme assumes the same set of events), we can then get a conflict.  
The degree of conflict depends on the temperament of the participants and 
on the surroundings.  Similarly, if people find that they have similar opinions about 
a “sick” problem, they get a good relationship and this relationship can be a basis
 for the future of the friendship (if one does not exist).  Going back to our r
estaurant problem, in the case of conflict the waiter has to make an extra-regular action:
 call the manager or police, and so on.
\par We are now ready to make a mathematical model of the conversation (Kovchegov 2004).
\vskip .2in
\par { \bf 2. A MODEL OF CONVERSATION AS A RANDOM WALK ON THE SEMANTIC TREE. }
\vskip .1in
\par This part contains information about the mathematical stochastic model of 
conversation and conflict. This model was realized as a program and the author obtains
 the same numerical results as well.  Since the full description of the model is rather 
lengthy, we present here only a toy version of the model.
\vskip .1in
\par { \bf A.	COMMON INTERESTS AND THREE LEVELS OF COMMUNICATION }
\vskip .1in
\par We first have to define what common interests mean in a given model.  If we have two 
words a and b and there exists a common prefix c, then a=c* and b=c*, where symbol * means 
the same set of letters (for an arbitrary alphabet).  If person one’s favorite set contains a and 
does not contain b and the second person’s favorite set b and does not contain a, we can say 
that person a has common interests on level c or c-interest.
\par We will differentiate three levels of communications: level one or the general level, 
level two or real common interests, when people are emotionally involved in the same 
business, and level three or private, heart-to-heart communications.  Second level communication 
generates (sometimes) deep positive or negative relations.  These types of communications 
are the basis for the emergence of new groups and the crash of old ones.  The third level of 
communication leads to love affairs and/or the emergence or destruction of families.  For this 
type of communication partners first use body language, the language of glances and touch.
\vskip .1in
\par { \bf B.	THE FREE VERBAL AND NONVERBAL COMMUNICATION MODEL  }
\vskip .1in
The free communication model is the model of a situation when participants do not have an 
agenda. We describe two types of free communication models.  The first type is the situation 
where participants sit in a given place. The second situation is the model of completely free 
communication, where participants do not have an agenda and move freely. The full 
description of the free conversation model for the case where people stay at a given 
location involves a more sophisticate mathematical language.
 \par    We start our description from the first type of communication model.  
Suppose we have a set of people I and a neighbor function N.  The set of subsets 
of I is called a partition of I if the union of both sets is the set I and the intersection 
of the two is the empty set. We call a partition of I a partition agreed with a 
neighbor’s function N if every element of the partition is a proper subset of at
least one N(j) for the same j. Let us denote the set partition of I that agrees with a 
neighbor’s function N by symbol H(I).  We can similar find the conditional probability for 
theme t(j,k) to be accepted as the suitable theme for conversation given as:
\par P{t(j,k)$\vert$j} =v ($\vert$sub(t(j,k),N(j))$\vert$ - 1)w(j,k)/$\vert$F(j)$\vert$,
where theme t(j,k) belongs to F(j), has number k, and $\vert$F(j)$\vert$ is the number of elements in set F(j).  
Symbol sub(t(j,k), N(j)) is the set of people from set N(J) that support theme t(j,k).  
We will later explain how to find the coefficient of v.  We need to, however, define
the random process for the set of marking partitions agreed with neighbor function N.  
Any element of the partition will be marked by theme type and time.  We use symbol Null 
to represent silence.  Like the theme of “Weather”, Null (or Silence) is a universal neutral theme. 
When a person stays silent it means that they are doing something: breathing, listening, eating, 
chewing, swallowing, watching, looking, sniffing, moving something, writing, reading, 
sleeping, walking, running, thinking, doing a job, and so on.  Coefficient v must be 
found from the condition: sum of all P{t$\vert$j}for all themes t from F(j) and Null equals one.
\par So if we want additional details we have to substitute actions instead of Null. When 
a local conversation for the same subgroup (element of partition) is over, all members get 
the same theme of Null and start a new time.  Everyone in this group became a subset
and new element of the partition. For instance, we have a conversation subgroup
B=$\{$1, 5, 9 $\}$ marked by theme b and time z.  Suppose, that when the time of conversation 
equals 23 minutes the conversation ends.  We need to transform subgroup B into three 
subgroups $\{$1$\}$, $\{$2$\}$, {9} and mark all three by theme Null.  The new local time must be started, too.
\par { \bf An example.}  Suppose we have 
I=$\{$1, 2, 3, 4 $\}$ and N(1)=$\{$ 1, 2, 4 $\}$, N(2)=$\{$ 2, 1, 4 $\}$, N(3)=$\{$ 3, 2, 4 $\}$, N(4)=$\{$4, 1, 3 $\}$. 
 Let us assume we have the second level communication and suppose that the favorite set
of themes F(j) has an empty intersection for all sets and a nonempty intersection only 
for every distinct two.  In this case partition A =$\{$ $\{$ 1, 2 $\}$, $\{$ 4, 3 $\}\}$ is 
the partition agreed with a neighbor’s function N. The marked partition is partition 
Am = $\{$ $\{$ 1, 2 $\}$, t(1,k); $\{$ 4, 3 $\}$, t(3,s) $\}$, where t(1,k) belongs to F(1) and 
t(3,s) belongs to F(3).  This means that th first conversation subgroup $\{$1, 2 $\}$ discusses 
theme t(1,k) and the second subgroup $\{$ 4, 3 $\}$ discusses theme t(3,s).  Let us infer that in 
random time z the second team stops talking, but the first team stays talking.  This signifies
that the initial marked partition will be transformed into the next one:
$\{\{$1, 2 $\}$, t(1,k); $\{$ 4, 3 $\}$, t(3,s) $\}$  $\rightarrow$  $\{\{$ 1, 2 $\}$, t(1,k); $\{$ 4 $\}$, Null; $\{$ 3 $\}$,Null $\}$.
\par Then there are few variants:
 $\{$ $\{$ 1, 2 $\}$, t(1,k); $\{$ 4, 3 $\}$, t(3,s) $\}$  $\rightarrow$  $\{$ $\{$ 1, 2 $\}$, t(1,k); $\{$ 4 $\}$, Null; $\{$ 3 4 $\}$,Null $\}$
 $\rightarrow$  $\{$ $\{$ 1 $\}$, Null; $\{$ 2 $\}$, Null; $\{$ 4 $\}$, Null; $\{$ 3 $\}$, Null $\}$  $\rightarrow$ 
 $\{$ $\{$ 1, 4 $\}$, t(4,k); $\{$ 2 $\}$, Null; $\{$ 3 $\}$,Null $\}$ and so on.  We can not use, for instance, partition 
$\{$ $\{$ 1, 3 $\}$, $\{$ 2, 4 $\}$ $\}$ because this partition does not agree with the neighbor’s function N (element 
$\{$1, 3$\}$ is not a part of N(j), for all j).  Nevertheless, in this case it is easy to describe all set H(I) 
(the set partition of I that agrees with neighbor function N):
H(I) = $\{$($\{$ 1 $\}$, $\{$ 2 $\}$, $\{$ 3 $\}$, $\{$ 4 $\}$), ($\{$ 1, 2 $\}$, $\{$ 3, 4 $\}$), ($\{$ 1, 2 $\}$, $\{$ 3 $\}$, $\{$ 4 $\}$), 
($\{$ 1, 4 $\}$, $\{$ 3, 2 $\}$), 
($\{$ 1, 4 $\}$, $\{$ 3 $\}$, $\{$ 2 $\}$), ($\{$ 1, 2, 4 $\}$, $\{$ 3 $\}$), ($\{$ 2, 3, 4 $\}$, $\{$ 1 $\}$), ($\{$ 1, 3, 4 $\}$, 
$\{$ 2 $\}$), ($\{$ 2, 3, 1 $\}$, $\{$ 4 $\}$) $\}$. 
   \par We thus have only 11 states and all of them are not fit for modeling the second level of conversation.  
Instance states ($\{$1, 2, 4$\}$, $\{$3$\}$), ($\{$2, 3, 4$\}$, $\{$1$\}$), ($\{$1, 3, 4$\}$, $\{$2$\}$), 
($\{$2, 3, 1$\}$, $\{$4$\}$) 
can be fit for level one conversation. 
 We deduce from our assumption: any three people
do not have common interests.  So for the modeling of solely the second level of communication 
we have to reject these partitions.  However, there does not exist pure communications in real life. 
More often than not real communication is a mix of all three levels. This means that free 
communication generates a process that emerges/clashes groups, families, and so on.  
What does this transition mean?  The transition 
($\{$1, 2$\}$, t(1,k); $\{$3$\}$, Null;  $\{$4$\}$, Null) $\rightarrow$ ($\{$1, 2, 3$\}$, t(1,s); $\{$4$\}$, Null) 
means that person 3 joins conversation team $\{$1, 2$\}$. Why is this possible?  
This occurs because team $\{$1, 2$\}$ changes its type of theme: they skip theme 
t(1,k)=crd and discuss theme t(1,s)=c, where c is level one’s theme and c, r, and d are words 
in a theme’s alphabet.  In this model the semantic field/scale contains alphabets (general letters, 
special letters) and sets of words. General letters are letters that code the general themes 
(the first level of communication).  A set of words has the structure (general words, special words) 
and grammar.  For example, “letter” Rm would code the general theme “Rumor” and “letter” Sp 
represents the theme “Sport.”  It then follows that word RmSp represents the theme: a rumor in sport.  
If we use the theme “Football,” “letter” Fo codes this and theme “Basketball” is coded by “letter” Ba 
and words SpFo represent football. Suppose that “letter” Gm represents theme “Game.”  We can 
find at least two people who like to talk about sport, but differentiate on the type of sport. So, if 
t(1,k)=SpFoGm and t({1, 2}, s)=RmSpFo, we can interpret this transition as the transition for 
the theme of talk ranging from questions about football to rumors about players, coaches, and so on.  
In this case the special word is transformed into the general word by adding one “letter” Rm.  
But rumors aside, only one letter transfers the special theme (special word) into the general 
theme (general word).  We define the special word as a word in a special alphabet.  
The general word is a word with at least one letter from a general alphabet.
This kind of explanation suggests to us that transition 
($\{$1, 2$\}$, t(1,k); $\{$3$\}$, Null;  $\{$4$\}$, Null) $\rightarrow$ ($\{$1, 2$\}$, t(1,s); $\{$3$\}$, Null;  $\{$4$\}$, Null)  
occurs before transition ($\{$1, 2$\}$, t(1,k); $\{$3$\}$, Null;  {4}, Null) $\rightarrow$ ($\{$1, 2, 3$\}$, t(1,s); $\{$4$\}$ , Null). 
 Finally, we have the chain of transitions ($\{$1, 2 $\}$, t(1,k); $\{$3$\}$, Null;  $\{$4$\}$, Null) $\rightarrow$ ($\{$1, 2$\}$, t(1,s); $\{$3$\}$, Null;  $\{$4$\}$, Null) )
$\rightarrow$ ($\{$1, 2, 3$\}$, t(1,s); $\{$4$\}$, Null).
\par We must only be concerned about the moment in time when there emerges a new 
theme that changes the structure of interests or sours future alterations.  If we want 
to create a completely free communication model where participants do not have an 
agenda and are able to move about, we have to use as a state the arbitrary partition of I.
\par The system then stays in some state (partition agreed with a neighbor’s function N) 
for a random period of time and can randomly change its state. If the time between 
transitions has an exponential distribution we have a Poisson process.  Let us define 
the probability of emergence of conversation subgroup A for a small period of time T
by formula Pr{emergence of conversation subgroup A} = m(A)T + o(T) where A is a 
subset of N(j), for the same j, $\vert$ A $\vert$ $\ge$  2 and the probability equals zero otherwise.  
Then if the intersection of all favorite subsets F for people from A is empty the theme for 
conversation is gotten from the general set G (level one conversation).  Otherwise a person
of subset A chooses the conversation theme from the intersection.  The probability of
collapse for a small period of time T equals M(A)T + o(T).  So the probability for conversation 
group A to emerge with theme t equals (m(A)T + o(T))(q(j)p(t|j) + $\ldots$ + q(k)p(t|k)),
where q(j) is the probability that person j initiates the conversation theme, p(t|j) is the 
conditional probability that theme t is chosen by person j and A=$\{$j, $\ldots$ ,k $\}$.
\par For the case of four people we do not have a large number of states or partitions.
  The first partition agrees with a neighbor’s function N (and
N(1)=$\{$1, 2, 4 $\}$, N(2)=$\{$2, 1, 4 $\}$, N(3)=$\{$3, 2, 4$\}$, N(4)=$\{$4, 1, 3$\}$) is D=($\{$1$\}$,Null; $\{$2$\}$, Null; $\{$3$\}$, Null; $\{$4$\}$, Null). 
The second group of marked partitions are partitions 
D(1,2;t)= ($\{$1, 2$\}$, t; $\{$3$\}$, Null; $\{$4$\}$, Null), D(1,3;t)= ($\{$1, 3$\}$, t; $\{$1$\}$, Null; $\{$3$\}$, Null), 
D(2,3;t)= ($\{$3, 2$\}$, t; $\{$1$\}$, Null; $\{$4$\}$, Null), D(1,4;t)= ($\{$1, 4 $\}$, t; $\{$3$\}$, Null; $\{$2$\}$, Null), 
D(2,4;t)= ($\{$4, 2$\}$, t; $\{$3$\}$, Null; $\{$1$\}$, Null), D(3,4;t)= ($\{$3, 4$\}$, t; $\{$1$\}$, Null; $\{$2$\}$, Null),
where conversation theme t belongs to the participants 
intersection of their favorite set or to the set of general  conversation themes.  
The third group of state is the group of partitions D(k,j;t1,t2) = ($\{$k, j$\}$, t1; $\{$s, r$\}$, t2), 
for all different combinations of $\{$k, j, s, r$\}$, where k, j, s, r belongs to the set 
I=$\{$1, 2, 3, 4 $\}$ and $\{$ k, j $\}$=$\{$ j, k,$\}$, $\{$r, s $\}$=$\{$s, r$\}$, and t1 and t2 are  the conversation themes.  
The last group of marked partitions is the group of partitions
D(j;t) = ($\{$k, s, r$\}$, t; $\{$j$\}$, Null), for different j, where  $\{$k, s, r$\}$ = I \ $\{$j$\}$, and t is the theme.
Symbol Null does not represents spiking actions (for instance, not verbal communication actions). 
\par We are now ready to write the equation for the first type of free communication model under 
one assumption: the time between transition has exponential distribution.  It may not be a realistic 
assumption (we achieve a pure Markov process), but we can at least finish the solution successfully.  
In the case of a semi-Markov process we get a lot of problems with analysis.  
\par If we do not care about themes we get the system of thirteen differential equations.  
We have a set of favorite themes F(j), where j belongs to I, and a set of general themes 
G (every person j may have the favorite set of general themes G(j) ).  In this case we get
1 + 3($\vert$F(1) F(2)$\vert$ + $\vert$F(1) F(3)$\vert$  + $\vert$F(1) F(4)$\vert$ + $\vert$F(2) F(3)$\vert$ + $
\vert$F(2) F(4)$\vert$ +  $\vert$F(3) F(4)$\vert$)  +  ($\vert$F(1) F(2) F(3)$\vert$ + ($\vert$F(1) F(2) F(3)$\vert$ + 
($\vert$F(1) F(2) F(3)$\vert$ +  ($\vert$F(1)F(2)F(3)$\vert$) + 16$\vert$G$\vert$ differential equations.
\par We have divided the communication process on the two independent parts.  
The first part is dedicated to process that emerges the communication groups 
(having verbal conversation groups first).  The second part of the communication 
problem is the stochastic model of theme flow in every subgroup, when the 
communication groups are formed. 
 
\par Step 1. We ignore the type of theme discussed. Let
D1 = D, D2 = D(1;T) = ($\{$1$\}$, Null; $\{$2, 3, 4 $\}$), D3 =  D(2;t),
 D4 = D(3,t), D5 = D(4,t), D6 = D(1,2;t), D7 = D(3,4; t), D8 = D(1, 4; t), 
D9 = D(2,3;t), D10 = D(2,4; t), D11 = D(1, 3; t), D12 = D(1, 2;t1,t2),
D13 = D(1, 3; t1, t2), D14 = D(1,4; t1, t2),  for all themes t, t1, t2.  
Let X(T) be our stochastic process and $p_j$(T) = Pr$\{$X(T) = Dj $\}$, where j= 1, $\ldots$ ,14.   
The system’s differential equation that describes the evolution of initial distribution is:

\par d$p_1$(T)/dT=(- m(1,2) - $\ldots$ - m(1,11))$p_1$ + m(2,1)$p_2$ + $\ldots$ + m(11,1)$p_{11}$ 
\par d$p_2$(T)/dT = -m(2,1)$p_2$ + m(1,2)$p_1$
\par d$p_3$(T)/dT = -m(3,1)$p_3$ + m(1,3)$p_1$
\par d$p_4$(T)/dT = -m(4,1)$p_4$ + m(1,4)$p_1$
\par d$p_5$(T)/dT = -m(5,1)$p_5$ + m(1,5)$p_5$
\par d$p_6$(T)/dT = -(m(6,1) + m(6,12))$p_6$ +m(12,6)$p_{12}$ + m(1,6)$p_1$
\par d$p_7$(T)/dT = -(m(7,1) + m(7,12))$p_7$ +m(12,7)$p_{12}$ + m(1,7)$p_1$
\par d$p_8$(T)/dT = -(m(8,1) + m(8,13))$p_8$ +m(13,8)$p_{13}$ + m(1,8)$p_1$
\par d$p_9$(T)/dT = -(m(9,1) + m(9,13))$p_9$ +m(13,9)$p_{13}$ + m(1,9)$p_1$
\par d$p_{10}$(T)/dT = -(m(10,1) + m(10,14))$p_{10}$ +m(14,10)$p_{14}$ + m(1,10)$p_1$
\par d$p_{11}$(T)/dT = -(m(11,1) + m(11,14))$p_{11}$ +m(14,11)$p_{14}$ + m(1,11)$p_1$
\par d$p_{12}$(T)/dT = -(m(12,6) + m(12,7))$p_{12}$ +m(6,12)$p_6$+m(7,12)$p_7$
\par d$p_{13}$(T)/dT = -(m(13,8) + m(13,9))$p_{13}$ +m(8,13)$p_8$+m(9,13)$p_9$
\par d$p_{14}$(T)/dT = -(m(14,10) + m(14,11))$p_{14}$ +m(10,14)$p_{10}$+m(11,14)$p_{11}$

where Pr$\{$ X(T+s) = Dk /X(T)=Dj $\}$=m(j,k)s + o(s),  for all j and k.
It is easy to find the stationary points (measure) of the system. Let put $m_{i,j}$ = m(i,j) for all i and j. 
\par First of all we will find eigenvector $ \bf v_1 $ for eigenvalue $ \lambda_1 =0 $. 
We will show that all solutions are going to $ \bf v_1 $, when time is going to plus infinity.
 How easy to find $ \bf v_1 $ = $ C_1(f_1, f_2, \ldots , f_5,f_6, \ldots , f_{11}, f_{12}, f_{13}, f_{14})^t$, 
where $f_1$ = 1, $f_2$ = $m_{1,2}$/$ m_{2,1}$, $f_3$ = $m_{1,3}$/$ m_{3,1}$, $f_4$ =
 $m_{1,4}$/$ m_{4,1}$, $f_5$ = $m_{1,5}$/$ m_{5,1}$, 
\par $f_{12}$ = $ (m_{7,12}m_{1,7}(m_{6,1}+m_{6,12}) 
+ m_{1,6}m_{6,12}(m_{7,1}+m_{7,12}))/(m_{7,1}m_{12,7}(m_{6,1}+m_{6,12}) + 
m_{6,1}m_{12,6}(m_{7,1}+ m_{7,12}))$,
\par $f_{13}$ = $ (m_{9,13}m_{1,9}(m_{8,1}+m_{8,13}) + m_{1,8}m_{8,13}(m_{9,1}+m_{9,13}))/
( m_{9,1}m_{13,9}(m_{8,1}+m_{8,13}) + m_{8,1}m_{13,8}(m_{9,1}+m_{9,13}))$, 
\par $f_{14}$ = $ (m_{11,14}m_{1,11}(m_{10,1}+m_{10,12}) + m_{1,10}m_{10,14}(m_{11,1}+m_{11,14}))/
( m_{7,1}m_{12,7}(m_{10,1}+m_{10,12}) + m_{10,1}m_{14,10}(m_{11,1}+m_{11,14}))$,
\par  $f_6$ = $(m_{12,6}f_{12} + m_{1,6})/(m_{6,1}+m_{6,12})$, 
$f_7$ = $(m_{12,7}f_{12} + m_{1,7})/(m_{7,1}+m_{7,12})$, 
$f_8$ = $(m_{13,8}f_{13} + m_{1,8})/(m_{8,1}+m_{8,13})$, 
$f_9$ = $(m_{13,9}f_{13} + m_{1,9})/(m_{9,1}+m_{9,13})$, 
$f_{10}$ = $(m_{14,10}f_{14} + m_{1,10})/(m_{10,1}+m_{10,14})$, 
$f_{11}$ = $(m_{14,11}f_{14} + m_{1,11})/(m_{11,1}+m_{11,14})$.
 $C_1$ = $ (f_1 + f_2+ \ldots + f_{14})^{-1}$.
\par Note. For checking the condition $(m_{1,2} + \ldots + m_{1,11})p_1$ = $m_{2,1}p_2 +
\ldots + m_{11,1}p_{11}$ we have to use identity for $f_{12}$: $ m_{1,6}m_{6,12}/(m_{6,1}+m_{6,12}) –
 f_{12} m_{6,1}m_{12,6}/(m_{7,1}+m_{7,12}) + m_{7,12}m_{1,7}/(m_{7,1}+m_{7,12}) –
 f_{12}m_{7,1}m_{7,12}/(m_{7,1}+m_{7,12}) = 0$. And similar identities for $f_{13}$ and for $f_{14}$.
We can represent the system of equation in the matrix form as: 
d${\bf  p}$(T)/dT =${\bf Ap}$(T),
where ${\bf p}$(T) is vector of probabilities and  {\bf A} is a matrix of coefficients.  In the case where all coefficients of intensity m(k,j) =1, for all k and j, where a is a positive constant our system of equation looks like: 
$ d{\bf p}(T)/dT = {\bf Ap}(T) $,
where matrix
 
${\bf  A}$= $ \left( \begin{array}{cccccccccccccc}
-10 & 1 & 1 & 1 & 1 & 1 & 1 & 1 & 1 & 1 & 1 & 0 & 0 & 0 \\
  1 & -1 & 0 & 0 & 0 & 0 & 0 & 0 & 0 & 0 & 0 & 0 & 0 & 0 \\ 
  1 & 0 & -1 & 0 & 0 & 0 & 0 & 0 & 0 & 0 & 0 & 0 & 0 & 0 \\
  1 & 0 & 0 & -1 & 0 & 0 & 0 & 0 & 0 & 0 & 0 & 0 & 0 & 0 \\
  1 & 0 &  0 & 0 & -1 & 0 & 0 & 0 & 0 & 0 & 0 & 0 & 0 & 0 \\
  1 & 0 & 0 & 0 & 0 & -2 & 0 & 0 & 0 & 0 & 0 & 1 & 0 & 0 \\
  1 & 0 & 0 & 0 & 0 & 0 & -2 & 0 & 0 & 0 & 0 & 1 &  0 & 0 \\
  1 & 0 & 0 & 0 & 0 & 0 & 0 & -2 & 0 & 0 & 0 & 0 & 1 & 0 \\
  1 & 0 & 0 & 0 & 0 & 0 & 0 & 0 & -2 & 0 & 0 & 0 & 1 &,0 \\
  1 & 0 & 0 & 0 & 0 & 0 & 0 & 0 & 0 & -2 & 0 & 0 & 0 & 1 \\
  1 & 0 & 0 & 0 & 0 &  0 & 0 & 0 & 0 & 0 & -2 & 0 & 0 & 1 \\
  0 & 0 & 0 & 0 & 0 & 1 & 1 & 0 & 0 & 0 & 0 & -2 & 0 & 0 \\
  0 & 0 & 0 & 0 & 0 & 0 & 0 & 1 & 1 & 0 & 0 & 0 & -2 & 0 \\
  0 & 0 & 0 & 0 & 0 & 0 & 0 & 0 & 0 & 1 & 1 & 0 & 0 & -2
 \end{array} \right) $
\par We can now find the exact solution of the system.  For this purpose we will find the eigen values for matrix$\bf  A$.  
\par The characteristic polynomial is  $\Delta(\lambda) $ = det$\vert$ ${\bf A}$ - $\lambda$ ${\bf I}$ $\vert$ = $\lambda$($\lambda^3$ + 15$\lambda^2$ + 46$\lambda$ + 28)
($\lambda^2$ + 4$\lambda$ +2$)^2$ ($\lambda$ + 1$)^3$ ($\lambda$ + 2$)^3$. 
\par The eigenvalues are $\lambda_1$ = 0, $\lambda_2$ = $\lambda_3$ = $\lambda_4$ = -1, $\lambda_5$ =  $\lambda_6$ = $\lambda_7$ = -2,     $\lambda_8$ = $\lambda_9$ = -2 -$\sqrt{2}$, $\lambda_{10}$ =  $\lambda_{11}$ = -2 +$\sqrt{2}$,
 $\lambda_{12}$ = z/3 + 29/z - 5   ,  $\lambda_{13}$ = - z/6 - 29/2z -5 + 1/($\sqrt{3}$(z/2 - 29/2z))  ,  
$\lambda_{14}$ = - z/6 - 29/2z -5 + 1/($\sqrt{3}$(z/2 - 29/2z)), where z = $( -648 + 3/\sqrt{26511})^{1/3}$.
\par For eigenvalue $\lambda_1$=0 we have one eigenvector $ {\bf  u_1 }$ = (1,1, $\ldots$ , 1$)^t$, for eigenvalue $\lambda_5$ = $\lambda_6$ = $\lambda_7$ = -1 we have three independing eigenvectors $ {\bf u_2 }$ = (0,0,1,-1, 0, $\ldots$ , 0$)^t$, $ {\bf u_3 }$ = (0,0,1,0, -1,0,  $\ldots$ , 0$)^t$, and $ {\bf u_4 }$ = (0,0,1,0,0,-1,0,  $\ldots$ , 1$)^t$.
\par For eigenvalue $\lambda_5$ = -2 multiple three we have three eigenvectors too: $ {\bf u_5 }$ = (0, 0, 0, 0, 0, 1,-1, 0, $\ldots$ , 0$)^t$, $ {\bf u_6 }$ = (0, 0, 0, 0, 0, 0, 0,1,-1, 0,  $\ldots$ , 0$)^t$
and $ {\bf u_7 }$ = (0, 0, 0, 0, 0, 0, 0, 0,1,-1, $\ldots$ , 0$)^t$. 
\par For eigenvalue $\lambda_8$ = $\lambda_9$ = -2-$\sqrt{2}$ we have two eigenvectors $ {\bf u_8 }$ = (0, 0, 0, 0, 0, -1/$\sqrt{2}$,-1/$\sqrt{2}$,0, 0, 1/$\sqrt{2}$, 1/$\sqrt{2}$, 1, 0,  -1 $)^t$ ,
${\bf u_9}$ = (0, 0, 0, 0, 0, 0, 0, -1/$\sqrt{2}$,-1/$\sqrt{2}$,1/$\sqrt{2}$, 1/$\sqrt{2}$, 0, 1, - 1$)^t$.
\par  For eigenvalue  $\lambda_{10}$ =  $\lambda_{11}$  = -2 + $\sqrt{2}$ we get two eigenvectors too: 
 $ {\bf u_{10}}$ = (0, 0, 0, 0, 0, 1/$\sqrt{2}$,1/$\sqrt{2}$,0, 0, -1/$\sqrt{2}$, -1/$\sqrt{2}$, 1, 0,  -1$)^t$ ,
${\bf u_{11}}$ = (0, 0, 0, 0, 0, 0, 0, 1/$\sqrt{2}$, 1/$\sqrt{2}$, -1/$\sqrt{2}$, -1/$\sqrt{2}$, 0, 1, - 1$)^t$.  
\par For eigenvalues $\lambda$ that satisfy equation $\lambda^3$ + 15$\lambda^2$ + 46$\lambda$ + 28 = 0 ( $\lambda_{12}$  ,  $\lambda_{13}$ ,  $\lambda_{14}$ ) eigenvector
is (1, $\frac{1}{1+\lambda}$, $\frac{1}{1+\lambda}$, $\frac{1}{1+\lambda}$, $\frac{1}{1+\lambda}$, $\frac{1}{1+\lambda}$, 
$\frac{1 +\lambda}{\lambda^2 + 4\lambda + 2}$, $\ldots$ , $\frac{1 +\lambda}{\lambda^2 + 4\lambda + 2}$, 
$\frac{2}{\lambda^2 + 4\lambda + 2}$,$\frac{2}{\lambda^2 + 4\lambda + 2}$,$\frac{2}{\lambda^2 + 4\lambda + 2}$,
$\frac{2}{\lambda^2 + 4\lambda + 2}$ $)^t$. We will receive the three eigenvectors $\bf {u_{12}}$,$\bf {u_{13}}$, $\bf {u_{14}}$ when we substitute 
three roots ($\lambda_{12}$=-11.075, $\lambda_{13}$ = -3.1132, $\lambda_{14}$=-0.81212)  in general formula.  
\par The sum of coordinates for eigenvectors ${\bf u_2}$, $\ldots$ , ${\bf u_{14}}$ equal zero:
$ \sum_{i=2}^{14} u_{k,i}$ = 0 for k=2, $\ldots$ , 14.    
\par  The fundamental matrix for our system is
\par  ${\bf X}$ (t)  = $\lbrack$  $C_1e^{\lambda_1t}{\bf u_1}$, $C_2e^{\lambda_2t}{\bf u_2}$,  $\ldots$ , $C_{14}e^{\lambda_{14}t}{\bf u_{14}}$ $\rbrack$.
\par The general solution of our system is ${\bf p}$(t,$C_1$,$C_2$, $\ldots$ , $C_{14}$) =  $C_1e^{\lambda_1t}{\bf u_1}$ + $C_2e^{\lambda_2t}{\bf u_2}$ +
  $\ldots$  +  $C_{14}e^{\lambda_{14}t}{\bf u_{14}}$, where $C_1$, $C_2$, $\ldots$ , $C_{14}$  are constants and
$\lambda_1$ = 0, $\lambda_2<  0$, $\ldots$ , $\lambda_{14} <  0$. So, when t $\rightarrow$  + $\infty$ the general solution going to $C_1$${\bf u_1}$.
 \par Let put $C_1$ =1/14 then the probability measure $C_1$${\bf u_1}$ =(1/14, $\ldots$, 1/14 $)^t$ will be only attractor for all solutions.
\par Thus, in the case of a free conversation (no preferred themes) after a short time the conversation is equally likely 
to switch to any other topic, including a stop, regardless of the initial distribution.

\par So, our system of linear equations (*) has the solution (see, for instance, [30]) 

p(T) = exp(e(1)T)Z1 + exp(e(2)T)Z2 + exp(e(3)T)Z3 + exp(e(4)T)Z4 + exp(e(5)T)Z5
+ exp(e(6)T)Z6 + exp(e(7)T)Z7 + exp(e(8)T)Z8 + exp(e(9)T)Z9 + exp(e(10)T)Z10
+ exp(e(11)T)Z11 + exp(e(12)T)Z12 + exp(e(13)T)Z13 + exp(e(14)T)Z14,

where matrices Z1, $\ldots$ , Z14 satisfy conditions ZiZj =0,I = Z1 + $\ldots$ +Z14, ZiZi = Zi , for all j and i from the set $\{$1, 2,$\ldots$ ,14 $\}$, I is identical matrix, and T is time

\par Step 2.  We can now describe the second part of the communication process.  For this, we need additional information about the person.
  We will then use the semantic field and body language. Suppose that A is the alphabet for the semantic field (the theme language).
The arbitrary word (every word represents a theme or a set of themes) is the sequence of letters.  For instance, a=S1H2K5 is word for
the alphabet A= {S1, H2, K5}. The conversation process builds or destroys random processes.  The typical trajectory of the random process
 looks like the sequence X(T0) =G2, X(T1) =AB, X(T2) = ABC, X(T3) = ABCD, X(T4)= ABCDF, X(T5) = G1, X(T6) =ABC, X(T7) = ABCKL, X(T8) =G3, etc.
 where T0, T1, T2, $\ldots$  are random moments in time.  This means that the conversation process starts from the general theme (G2) and 
proceeds onto special questions.   We can see a growing deepness in conversation (T1-T5) and at moment T5 the conversation process goes
 back to the general area again, and so on. How can we formalize this type of process?  We use the randomized formal grammar method.  What
 do mean by this?  Let us propose that the participants have a non-empty set of common favorite themes.  For instance, suppose that theme 
b= ABCDF is a common favorite theme for all members of the conversation team.  Someone initiates the following approach: she or he offers 
 theme AB.  Someone then makes the next approach (theme ABC) and so on.  In a few steps the conversation subgroup reaches the desirable 
theme ABCDF.  When participants tire or the theme is exhausted the process moves to the opposite direction.  Our next question is how do 
different evaluations and/or opinions about events (inside theme ABCDF) transform relations (see 3rd step). We represent the semantic scale 
as trees: S-trees for special topics, G-trees for general topics (funny stories, jokes, anecdotes, rumors, and C-trees for current events.  Suppose 
we the alphabet of themes A and suppose that any nodes on the tree are marked by letters from the alphabet and by a set of events.  The roots 
are marked by symbol S for S-tree, symbol G for G-tree, and symbol C for C-tree.  Now let us assume that all three symbols do not belong to the 
alphabet.  The nodes on the tree represent words.  How do we find these words?  We take the shortest path from the root to a given node and 
 write down (from left to right) all letters that lie on the path from the root to given node.  The favorite or sick sets of themes are the set of nodes
 on the G – or S – trees.  In this case the conversation process is a random walk on the trees. We can combine all trees into one by adding one 
additional extra root with three edges to the roots of S-, G-, and C- trees.  How do we calculate the probability of coming from a given node to 
the next neighbor or to jump to another place?  Suppose we have node j on the semantic tree.  For every participant we calculate the set of
 shortest paths from a given node to all elements of the favorite set.  For a given person we must calculate the number of paths that go from 
a given node j to a member of the favorite set with the nearest edges of node j (this means that the edge is a part of the trajectory).  The probability
 of going onto the next node must be proportional to its number.  For instance, if node j has the set of neighbors {k, l, r} on the semantic tree with 
a number of paths that start at j, that go throughout k (or l or r) equaling to M(j,k) (or M(j,l) or M(j,r)), this probability walk to k must be proportional
 M(j,k) and so on.  We can now demonstrate an example of a conversation model (random walk on the tree model).  Suppose we have alphabet A and
 a semantic tree $\{$ (S,Cu), (S,Sp), (S,Ar), (Cu,D1), (Cu,D2), (Sp,Fu), (Sp,Ba), (Ar,Mu), (Ar,Pa) $\}$
 for two people (having case one of six partitions D(1,2) = ({1, 2}, {3}, {4}), D(1,3), D(1,4), D(2,3), D(2,4), D(3,4)) and 
case D(2).The numeration of themes are 1 for S, 2 for Cu, 3 for Sp, 4 for Ar, 5 for D1, 6 for D2, 7 for Fu,
8 for Ba, 9 for Mu,  10 for Pa.
  The favorite sets are F1=$\{$Cu, D1, Sp, Ba$\}$ or $\{$2, 5, 3, 8$\}$ and F2=$\{$Sp, D2, Fu, Ba$\}$ or$\{$ 3, 6, 7, 8 $\}$. 
\par  This signifies that we will use a number theme instead of a word theme (i.e. number 2 instead of Cu (Cooking)).  The arrow and number on the 
arrow denotes the probabilities of skipping onto the next position.  We have the graph of transition and we can therefore write the system of 
differential equations. Let us denote the conditional probability Pr{X(T)=D(A,t) / X(T)=D(A)} by symbol p(t,A;T), where t represents the 
theme for conversation  group A (suppose A= $\{$1, 2$\}$ or $\{$1, 3$\}$ or $\{$1, 4 $\}$ or  $\{$ 2, 4$\}$).
\par The system’s differential equation that describes the evolution of the initial distribution on the semantic tree is: (Suppose D(A) is D(1,2) = ($\{$1, 2 $\}$, $\{$3$\}$, $\{$4$\}$,  F1 = $\{$2, 5, 3, 8$\}$,  F2= $\{$ 3, 6, 7, 3, 8 $\}$, 
Pr{X(T+s) = D(A,t) /X(T)=D(A)}=n(j,k)s + o(s), n(j,k) $\>$ 0  and let p(t,T) = p(t,$\{$1,2$\}$;T))

\par dp(1,T)/dT  = -(n(1,2) +  n(1,3) + n(1,4))p(1,T) + n(2,1)p(2,T)  + n(3,1)p(3,T) + n(4,1)p(4,T)
\par dp(2,T)/dT  = -(n(2,1) +  n(2,6))p(2,T) + n(1,2)p(1,T)  + n(5,2)p(5,T) + n(6,2)p(6,T)
\par dp(3,T)/dT  = -(n(3,1) +  n(3,8))p(3,T) + n(1,3)p(1,T)  + n(7,3)p(7,T) + n(8,3)p(8,T)
\par dp(4,T)/dT  = -(n(4,1) +  n(4,9))p(4,T) + n(1,4)p(1,T)  + n(9,4)p(9,T) + n(10,4)p(10,T)
\par dp(6,T)/dT  = -n(6,2)p(6,T) + n(2,6)p(2,T)
\par dp(8,T)/dT  = -n(8,3)p(8,T) + n(3,8)p(3,T)
\par dp(9,T)/dT  = -n(9,4)p(9,T) + n(4,9)p(4,T)
\par dp(5,T)/dT  = -n(5,2)p(5,T)
\par dp(7,T)/dT  = -n(7,3)p(7,T)
\par dp(10,T)/dT = -n(10,4)p(10,T)
\par p(1,T) + p(2,T) + p(3,T) + p(4,T) + p(5,T) + p(6,T) + p(7,T) + p(8,T) + p(9,T) + p(10,T) =1
Note: Let us not forget that numbers represent conversation themes.

\par It is easy to find the stationary points (measure p(k,T)=p(k), for all T) of the system and we can 
additionally find the system’s solution.  We need to write similar equations for all partitions and then 
plan (at a later point) to also use the “natural attractions.”  This attractiveness is a very important 
property and it known as the relational property.
\par We can similarly make the model of conflict.  The model of conflict pays more attention to details
and therefore essentially generates more states of the system and large numbers of equations.  
If we want additional details we can use body language as well.  For instance, in the third level of 
communication (greetings, heart-to-heart) body language is more important than oral.  But for all of 
these models we need to write a large system of equations.
\vskip .2in
\par { \bf 3.	THE INSURANCE COMPANY: A SEMANTIC REPRESENTATION OF INTERNAL FLOW AND A STATISTICAL SIMULATION OF THE COST OF A DISEASE.}
\vskip .1in
\par An insurance company is an example of a company in which the main business 
is almost purely informational. The real process (emerging diseases, interaction 
of patients and doctors, diagnosis, procedures, payments, etc.) is omitted from 
insurance company (IC).  These actions, however, or the majority of them need 
to be reflected by a system of informational flows.  An insurance company creates 
a few doctor networks, hospital networks, and forms the membership by selling 
insurance packages (“products”).  The product consists of a set of rules for the 
patients purchasing the package.  The cheaper package prescribes the member 
doctors, hospitals, and so on, while the more expensive package provides patients 
more freedom with additional choices.  The informational and reality images are 
not the same, and is the main reason for the search of fraudulence, auditing, and 
so on, all of which are an important part of IC activities.
\par Insurance company activities cover few areas consisting of: enrollment 
and maintenance (sell products to individual and group members), and providers 
(institutional and professional). Brokers provide new member enrollments, while 
vendors are responsible for doctor’s networks that want to work for given 
insurance companies. Following, diseases and/or monitoring force members 
to interact with doctors. These processes control the insurance company: 
from the first encounters with doctors, the IC gets claim/encounter information.
The claim contains diagnosis code, procedure’s code, billing information, patients’ name, 
and so on. The insurance company evaluates all this information by using some 
criterions and then makes payments or investigates the case. So, the two processes 
of selling of company’s products (health care insurance plans) and diseases or 
preventive monitoring are two base processes for an IC. Information and people 
that provide this support (programmers and report’s provider) support the two
 base processes. The computer and mathematical modeling of both external base 
processes is discussed by the author (Kovchegov, 2000). 
\par The main objects of IC are populations (IC want makes him members), 
network of doctors and hospitals, brokers, vendors and another functions 
typical for all companies (human resource, payroll department, fraud department and so on). 
The internal information process can be described through a linguistic method: 
using semantic trees (semantic code) and grammar structure. We can describe 
the main subject areas that almost cover the whole semantic space of an IC: 
Member (personal insurance), Group (employer insurance), Encounter/Claim, 
Product, Provide, Professional, Capitation, Pharmacy, Business entities, Finances, 
and Networks. The next levels are field names and tables. We can describe every 
table and documents by writing words using the subjects name, table and field language. 
So, the report can look like SA1 (f1, f4, f35), SA2 (f2, f8), f (22,11), f (2,57), 
where SA1 is the subject area “Member”, SA2 is “Encounter/Claim”, f1, f4, f35 and so on 
are base field names (field’s alphabet). Then we describe the set of possible functions: 
extractions, merging (“assemble”), creation of new fields (sum(f3,f7, $\ldots$ , f78)) and indexes 
and so on. So we can do a semantic tree for an IC and describe the grammar of the 
“reports’ language”.  Unfortunately it is not easy to find a training corpus that is 
necessary for getting a formal grammar of language for the report (and for another 
internal language). This is our main reason why we concentrate our attention on 
“doctor – patient” base process.
This section of the paper is dedicated on demonstrating the capability to represent the 
external flow in semantic form.  We create the alphabet (“procedure’s alphabet) and 
then find the formal grammar that generates patient diseases histories.  The formal 
grammar and frequencies give us the ability to calculate the cost of an arbitrary disease
by using the method simulation.  For this purpose we use two methods: the first called 
a “ direct statistical method” that uses some hierarchical system of calculation to do the 
calculation.  The second method, described later on, is divided in two parts.  The first 
part contains the calculations and extrapolation set of “normative variables” and 
“technological chains.”  The keystone element of the second part is the simulation of 
a price formation process. Our main goal, nonetheless, is to describe the alphabet 
and find the formal grammar.
\par We now describe the type of information used.  Patients that stay in
 hospitals are called inpatients and information on these patients’ is called 
institutional files. Other patients are called outpatients with an associated 
professional file. The five tables (see Appendix A) are an extraction from a 
professional file, years 1995-99, with the disease of diabetes.  We use the alphabets 
$\{$ A$\_$, AN, DR, E$\_$, LP, L$\_$, M$\_$, S1, S2, S3, S4, SO, RD, $\ldots$ $\}$, 
where “letter” A$\_$ stands for procedure “TRANSPORTATION,” “letter” AN 
stands for procedure “ANESTHESIA,” “letter” DR stands for procedure “DRUG,” 
“letter” LP stands for procedure “LAB PATOLOGY,” “letter” E$\_$ stands for procedure 
“DIGEST SYSTEM,” “letter” M$\_$ stands for procedure “MED  SERVICE,” “letter” 
S1 stands for procedure “SURGERY:INTEG,” letter S4 stands for procedure 
“SURGERY: CARD,” and so on.  For diabetes we use an alphabets of 37 letters.  
Every patient has a sequence of the procedures.  We have transferred the sequence 
of procedures into the set of “long” and “short” words.  The “long” word looks like 
"A$\_$-5AN-2DR-11 S4-1 S2-2", where a letter represents a procedure, a number after
 letter denotes the number of times.  So, A$\_$-5 means that procedure “TRANSPORTATION” 
is used five times. The “short” word is the “long” word without a number.
The “short” word looks like "A$\_$ANDR S4 S2.” We then calculate the 
frequencies of “short” words for all five years.
\par  We can see from the complete lists that the majority of words are small length. 
 The bigger “words” have a less probability of occurring.  The first step is to generate 
the list of words (randomly) and then calculate the price associated with every word. 
 For the real calculation we generate “words with frequencies”:  “AN-1; DR-17; S4-2; S5-1.”  
These words signify that the patient had anesthesia once, surgery (digest) – twice, surgery 
(cardiology) – once and took drugs 17 times.  Once the list of  “words with 
frequencies” has been generated, the program calculates the price of the list.
 \vskip .2in 
\par { \bf A.	THE SYNTACTIC MODEL AND THE FORECAST FOR THE SYNTACTIC MODEL.}
 \vskip .1in
\par We now transfer our “short words” problem into grammar problems.  This means that for 
a list of short words (see list for 1995–99) we generate this word’s automaton grammar. 
 Then, when we get the five grammars for all five tables, we find the general pattern for 
all stochastic grammar.   We thus reduce our problem into one: to make a forecast for the 
matrix of probability.  This means we have to make a prognosis for multidimensional 
numerical vectors.  In the general case there are a lot of solutions for this problem, but 
this method does not allow us to make a prognosis.  We start our analysis from year 1998, 
which contains more information.
\par { \bf Step 1. Grammar for table 4.}
\par We have grammar G4=(VN, VT, P, S), where S is the start or root symbol, P is the set of substitute 
or deduced rules, VT is the set of terms or set of words (having all letters belong to our alphabet), 
VN is the set of non terms, denoted as regular grammar.  This signifies that all deduced rules look 
like A $\rightarrow$ aB or A $\rightarrow$ a, where A and B are non-terms, and a is a word in the 
given alphabet.  We only need the description of set P.  Our goal is to obtain a regular grammar, 
which generates a given set of short words.  The set of deduced rules for words from table 4 is shown below.
S  $\rightarrow$  A$\_$, S   $\rightarrow$ A$\_$X1, X1  $\rightarrow$  ANX2, X2  $\rightarrow$   DRX3, 
X3 $\rightarrow$  LP, X3 $\rightarrow$ S4, X1  $\rightarrow$   E$\_$, X1   $\rightarrow$   DR, 
X1   $\rightarrow$  LP, X1  $\rightarrow$   DRX4,X4 $\rightarrow$ E$\_$X5, X6 $\rightarrow$   S4, 
X5  $\rightarrow$  LP, X4  $\rightarrow$ E$\_$, X4  $\rightarrow$  LP, X4   $\rightarrow$  S1,
X4 $\rightarrow$  S4, S  $\rightarrow$   DR, S  $\rightarrow$  DRX6, X6  $\rightarrow$   LPX7, 
X7  $\rightarrow$  RDX8, X7  $\rightarrow$  RD, X8  $\rightarrow$ S4, X7 $\rightarrow$ S4,
X7   $\rightarrow$ S1, X7  $\rightarrow$  S1X9, X9  $\rightarrow$  S4, X6  $\rightarrow$  RD,
 X6 $\rightarrow$  SO, X6   $\rightarrow$  S1, X6  $\rightarrow$ S1X7, X6  $\rightarrow$  S2, 
X6   $\rightarrow$  S4, S  $\rightarrow$   AN, S  $\rightarrow$  ANX10, X10   $\rightarrow$  DR, 
X11   $\rightarrow$  DRX12, X12  $\rightarrow$  S4.
\par We can minimize the number of substitutions and calculate the probability of using the 
given rule.  We obtain the number rule and the probability of using this rule of substitutions.  
In our grammar all words are generated from the table as well as additional information.  
But the situation is not that bad because the arbitrary table contains only partial information 
and short words, which have larger frequencies.   In actuality, short words can be very long.  
For larger words the probability of occurrence is less than the probability of a shorter one.

\par For future application we will represent the above grammar in a hierarchal or tree form: 
we divide all rules into levels. The first level is the root level and contains all substitution 
rules with first symbol S.  The second level rules are a set of rules in which the first 
symbol belongs to the previous set, the first level rules, and so on.
For our case the first level rules is:
S  $\rightarrow$  A$\_$, S  $\rightarrow$  DR, S   $\rightarrow$  AN, S   $\rightarrow$  E$\_$, 
S  $\rightarrow$  LP, S   $\rightarrow$   S1, S   $\rightarrow$  S4, S  $\rightarrow$  S4, S $\rightarrow$ RD.
\par The second level rules is:
A$\_$  $\rightarrow$  AN, A$\_$  $\rightarrow$  DR, A$\_$   $\rightarrow$  E$\_$, A$\_$  $\rightarrow$  LP,
A$\_$   $\rightarrow$  S1
DR $\rightarrow$ E$\_$, DR   $\rightarrow$  G$\_$, DR   $\rightarrow$  LP, DR  $\rightarrow$  RD, DR  $\rightarrow$  RD, 
DR   $\rightarrow$  S1, DR $\rightarrow$  S2, DR $\rightarrow$  S4, AN  $\rightarrow$  DR

\par The third level is the set:
DR $\rightarrow$  E$\_$, DR  $\rightarrow$  LP, DR  $\rightarrow$  S1, DR   $\rightarrow$  S4, AN   $\rightarrow$  DR, 
LP   $\rightarrow$  RD, LP  $\rightarrow$  S1, LP  $\rightarrow$  S4

\par The forth level contains the set of rules:
E$\_$  $\rightarrow$  LP, E$\_$  $\rightarrow$  S1, E$\_$   $\rightarrow$  S4
and so on.
\par In the future we use the following notation for rules.  The notation A$\rightarrow$ B for rules 
of level N signifies that (a) there does not exist a rule of level N+1 that begins with symbol B; 
(b) the calculation process prints out a word and goes to the beginning of the calculation process.  
The notation A $\rightarrow$  B* means that there exists a next level rule which starts from symbol B.  
We write A $\rightarrow$ B, A $\rightarrow$  B*, when we want say that there exists 
both a modification and a non-zero probability. 
\vskip .2in
\par {\bf Frequencies for Grammar 4 (1998) }
\vskip .2in
\par 1st level  (S is the starting point, followed by first level rules)
\vskip .1in
 \centerline{ \bf  S} 
 \begin{center} 
 \begin{tabular} { cccccccccccc }
A$\_$   &	A$\_$*  &  DR  &  DR*  &  AN  &  AN*  &  E$\_$  &   LP  &   LP*  &  	S1  &  	S4  &	RD  \\
.14348  &  	0.05219  &	.43117  &   .2581  &  .010523  &   .007252  &  .00384   &  .0506  &   .00597  &  	.01735  &   .01948  &   .0000  
\end{tabular}
 \end{center}
\vskip .1in
\par 2nd level (rules for the second level are used only for letters with asterisks: A$\_$*, DR*, AN*, and LP*)
\vskip .1in
 \centerline{ \bf  S  $\rightarrow$  A$\_$ }
\begin{center} 
 \begin{tabular} { cccccc  }
AN*  &   DR  &    DR*  &   E$\_$  &   	LP  &	S1  \\
0.1110  &  .2597  &	   .36769  &   .180376  &   	.015857  &   	.01189
\end{tabular}  
\end{center}  
\vskip .1in
  \centerline{ \bf  S  $\rightarrow$   DR}
 \begin{center} 
 \begin{tabular} { c c c c c c c c c c c  }
E$\_$  &  	G$\_$  &  G$\_$S4  &  LP  &   LP*  &	RD  &   SO  &   S1  &  S1S4  &   S2  &	S4  \\
.00716  &  	.00881  &   	.00551  &  	.30689  &   	.1901  &   .0099  &   .0066  &   .1041  &   .01267  & 	.0066 &   .34159
\end{tabular}  
\end{center} 
\vskip .1in
 \centerline{ \bf  S  $\rightarrow$ AN}
\vskip .1in
\begin{center} 
 \begin{tabular} { c c }
DR  &   DRS4   \\
0.78431  &  	0.21569
\end{tabular} 
\end{center} 
\vskip .1in
Note. In this paper we do not present events with small frequencies. Symbol LP* occurs with a small empirical probability 
(frequency equal 0.00597), so we ignore the next level.
\vskip .1in
\par { \bf 3rd level (use third level rules for AN*, DR*, LP*and ignore off events) }
\vskip .1in
\centerline{ \bf S  $\rightarrow$ A$\_$   $\rightarrow$  DR }
\begin{center} 
\begin{tabular}{llllll}
E$\_$	  &   E$\_$*  &   LP  &   LPS4  &   S1  &   	S4  \\
0.34837  &   	.18797  &   	.12030  &   .04762  &   .070175  &   .155388
\end{tabular}
\end{center} 
\vskip .1in
\centerline{ \bf  S  $\rightarrow$  A$\_$   $\rightarrow$  AN }
\begin{center}  
\begin{tabular} { c c }
DR  &   DR*  \\
.5089  &  .49107
\end{tabular} 
 \end{center} 

\vskip .1in
\centerline{ \bf  S  $\rightarrow$  DR $\rightarrow$ LP }
\begin{center} 
 \begin{tabular} { ccccc }
RD  &	RDS4  &  S1  &  S1S4  &  S4  \\
0.03478  &  	.02898  &  	.08985  &  	.03188  &	.8145
\end{tabular}  
\end{center} 
\vskip .1in
\par { \bf 4th level  (use fourth level rules for E$\_$*, DR* and ignore off events) }
\vskip .1in
\centerline{ \bf  S  $\rightarrow$ A$\_$  $\rightarrow$ DR $\rightarrow$ E$\_$  }
\begin{center} 
 \begin{tabular} { cccc }
LP  &	LPS4  &   S1  &  S4  \\
0.25333  &	0.2400  &	0.1333  &	0.37333
\end{tabular} 
\end{center} 
\par The grammar can now generate the list of short words that contain more words than included in table 4. We must, likewise, 
describe the grammar frequencies for all tables and then create a general universal grammar that generates all tables.  It is easy 
to demonstrate that the above-described grammar is our universal grammar.  What is the difference between these grammars? 
 The difference is just the Frequencies for Grammars and a forecast only needs to be done for this.  We should be aware that our
 frequencies are in numerical format.  We will show below the frequencies for all cases, only for the first two levels.  All frequencies 
below are frequencies for a universal grammar.  If the universal grammar with the given list of frequencies generates a set of 
words from table N, we call this: Frequencies for Grammar N.
\vskip .2in
\par { \bf  Frequencies for Grammar 1 (1995, Table 1 from Appendix A) }
\vskip .1in
\par  1st level
\vskip .1in
\centerline{ \bf S }
\begin{center} 
 \begin{tabular} { cccccccccccc }
A$\_$  &   A$\_$*  &  DR  &  DR*  &  AN  &   AN*  &	E$\_$  & 	LP  &	LP*  &   S1  &  S4  &   RD   \\
0.0281  &   0.0355  &   0.528  &  	 .1139  &   .01390  &  .025214  &   .00003  &  	.0754   &   .00007  &   .02515  &	  .0281  &  0.000
\end{tabular}  
\end{center} 

\vskip .1in
\par 2nd level
\vskip .1in

 \centerline{ \bf S $\rightarrow$ A$\_$ }
\begin{center} 
 \begin{tabular} { cccccc }
AN*  &   DR	&  DR*  &   E$\_$  &   LP  &  S1  \\
0.999  &  0.0001  &   0.0003  &  0.0003  &   0.0001  &   0.0002
\end{tabular} 
\end{center} 
\vskip .1in
 \centerline{ \bf  S   $\rightarrow$  DR }
\begin{center} 
 \begin{tabular} { ccccccccccc }
E$\_$  &    G$\_$  &   G$\_$S4 &	LP  &  	LP*  &   RD  &   SO  &   S1  &  S1S4  &   S2  &  	S4  \\
.00000  &   .0000  &  0.0000  &  	.4545  &   .0001  &   .0000  &   .0000  &   .3246  &   .0000  &   .0000  &  	0.2208
\end{tabular} 
\end{center}  

\vskip .2in
\par  {\bf Frequencies for Grammar 2 (1996, Table 2 from Appendix A)  }
\vskip .1in
\par 1st level
\vskip .1in
 \centerline{ \bf  S }
\begin{center} 
 \begin{tabular} { cccccccccccc }
A$\_$  &   A$\_$*  &   DR  &  DR*  &   AN   &    AN*  &     E$\_$  &     LP  &      LP*  &   S1  &   S4  &    RD  \\
0.0318  &   0.0710  &   0.4388  &   .1929  &   .044  &   .01325  &   	.0000  &   .0991  &	  .02915  &   .0079  &   .0668  &   0.0053
\end{tabular} 
\end{center} 

\vskip .1in
\par 2nd level
\vskip .1in
\centerline{ \bf  S $\rightarrow$ A$\_$  }
\begin{center} 
 \begin{tabular} { cccccc}
AN*  &  DR  &  DR*   &   E$\_$  &   LP  &   S1  \\
0.5149  &   0.1940  &   0.1940  &  0.0970  &   0.00004  &   0.00006  
\end{tabular}  
\end{center} 
\vskip .1in
 \centerline{ \bf S $\rightarrow$  DR  }
 \begin{center}
 \begin{tabular} { ccccccccccc}
E$\_$  &   G$\_$  &   G$\_$S4  & 	LP &   	LP*  &   RD  &   SO  &   S1  &   S1S4  &   	S2  &   S4  \\
.00000  &   .0000  &   0.0000  &   .3874  &   .1730  &   .0439  &   .0522  &   .0549  &   .0000  &   .0000  &   0.28846
\end{tabular}  
 \end{center} 
\vskip .2in
\par {\bf  Frequencies for Grammar 3 (1997, Table 3 from Appendix A) }
\vskip .1in
\par 1st level (AN* is ANDR, LP* is LPS4)
\vskip .1in
\centerline{ \bf S }
 \begin{tabular} { cccccccccccc}
A$\_$  &   A$\_$*  &   DR  &   DR*  &   AN  &  AN*  &   E$\_$  &   LP  &   LP*  &   S1  &   S4  &   RD  \\
.0515  &  0.1019  &   .3772  &   .1967  &   .0342  &   	.007252  &   	.00468  &   .1131  &   .0288  &   .0061  &   .0749  &   .0000
\end{tabular} 
\vskip .1in
\par 2nd level (LP* is LPS4)
\vskip .1in
 \centerline{ \bf S $\rightarrow$ A$\_$ }
 \begin{center} 
 \begin{tabular} { cccccc}
AN*  &   DR  &   DR*  &   E$\_$  &   LP  &   S1  \\
0.3533  &   .2827  &   .1943  &   .1696  &   .0000  &   .0000
\end{tabular} 
\end{center}  

\vskip .1in
\centerline{ \bf  S $\rightarrow$ DR}
\begin{center}
 \begin{tabular} { ccccccccccc}
E$\_$  &   G$\_$  &   G$\_$S4  &   LP  &   LP*  &   RD  &   SO  &   S1  &   S1S4  &   S2  &   S4  \\
.0000  &   .0000  &   .0000  &   .3681  &   .1831  &   .0403  &   .0201  &   	.0714  &   .0000  &   .0000  &   .3168
\end{tabular} 
 \end{center}

\vskip .2in
\par {\bf Frequencies for Grammar 5 (1999, Table 5 from Appendix A) }
\vskip .1in
\par {\bf 1st level  }
\vskip .1in
\centerline{ \bf  S }
\begin{center} 
 \begin{tabular} { cccccccccccc}
A$\_$  &   A$\_$*  &   DR  &   DR*  &   AN  &   	AN*  &   E$\_$  &   LP  &   LP*  &   S1  &   S4  &   RD  \\
0.0589  &   0.1406  &   .4470  &   .2856  &   .00001  &   .00001  &   .0000  &   .0333  &    .0000  &  .0208  &   .0136  &   	.0000  
\end{tabular} 
 \end{center} 
\vskip .1in
\par { \bf 2nd level (LP* is LPS4) }
\vskip  .1in
\centerline{ \bf S  $\rightarrow$  A$\_$ }
\begin{center} 
 \begin{tabular} { cccccc}
AN*  &   DR  &   DR*  &   E$\_$  &   LP  &   S1  \\
0.0000  &   .3898  &   .4550  &   .1552  &  .0000  &   .0000
\end{tabular} 
 \end{center} 
\vskip .1in
 \centerline{ \bf S $\rightarrow$ DR }
 \begin{center} 
 \begin{tabular} { ccccccccccc}
E$\_$  &   G$\_$  &   G$\_$S4  &   LP  &   LP*  &   RD  &   SO  &   S1  &   S1S4  &   S2  &   S4  \\
.0000  &   .0000  &   .0000  &   .2907  &   .1895  &   .0000  &  .0000  &   .1504  &   .0000  &   .0000  &   .3624
\end{tabular} 
 \end{center} 
\vskip .1in
\par We now have the information needed to make a forecast.  We gather information for the first level together (see below).
\vskip .1in
\par {\bf 1st level (1995-99) }
\begin{center} 
 \begin{tabular} { cccccccccccc}
A$\_$  &   A$\_$*  &   DR  &   DR*  &   AN  &   	AN*  &   E$\_$  &   LP  &   LP*  &   S1  &   S4  &   RD  \\
.0281  &  .0355  &   .528  &   .1139  &  .01390  &   .025214  &   .00003  &   .0754  &   .00007  &   .02515  &   .0281  &   	.000  \\
.0318  &   .0710  &   .4388  &   .1929  &   .044  &   .01325  &   .0000  &   	.0991  &   .02915  &   .0079  &   .0668  &   .0053  \\
.0515  &   .1019  &   .3772  &   .1967  &   .0342  &   .007252  &   .00468  &   .1131  &   .0288  &   .0061  &   .0749  &   .0000  \\
.14348  &    .05219  &   .43117  &   .2581  &   .010523  &   .007252  &   .00384  &   .0506  &   .00597  &   .01735  &   .01948  &   .0000  \\
.0589  &   .1406  &   .4470  &   .2856  &   .00001  &   .00001  &   .0000  &   .0333  &   .0000  &   .0208  &   .0136  &   .0000
\end{tabular} 
 \end{center} 

\par We have a difficult problem of predicting 12-dimensional vectors in an 11-dimensional space (sum of all elements in a row must equal one). 
 A similar problem exists for level two (5-dimensional and 10-dimensional).
\vskip .1in
\par {\bf 2nd level (1955-99) }
\vskip .1in
  \centerline{ \bf S $\rightarrow$ A$\_$ }
\begin{center} 
 \begin{tabular} { cccccc}
AN*  &   DR  &   DR*  &   E$\_$  &   LP  &   S1  \\
0.999   &   0.0001  &   0.0003  &   0.0003  &   0.0001  &   0.0002  \\
0.5149  &	0.1940  &   	0.1940  &   	0.0970  &   	0.00004  &   	0.00006  \\
0.3533  &	0.2827  &   	0.1943  &   	0.1696  &   	0.0000  &   	0.0000  \\
0.1110  &   	0.2597  &   	0.36769  &   	0.180376  &   0.015857  &   0.01189  \\
0.0000  &   	0.3898  &   	0.4550  &   	0.1552  &   	0.0000  &   	0.0000
\end{tabular} 
 \end{center} 
\vskip .1in
  \centerline{ \bf S $\rightarrow$ DR }
\begin{center} 
 \begin{tabular} { ccccccccccc}
E$\_$   &   G$\_$  &   G$\_$S4  &   LP  &   LP*  &   RD  &   SO  &   S1  &   S1S4  &   S2  &   S4  \\
.00000  &     	.0000  &   .0000  &   .4545  &   .0001  &   .0000  &   .0000  &   .3246  &   .0000  &   .0000  &   .2208  \\
.00000  &   	.0000  &   .0000  &   .3874  &   .1730  &   .0439  &   .0522  &   .0549  &   .0000  &   .0000  &   .28846  \\
.0000    &   	.0000  &   .0000  &   .3681  &   .1831  &   .0403  &   .0201  &   .0714  &   	.0000  &   .0000  &   .3168  \\
.00716  &   	.00881  &   .00551  &   .30689  &   .1901  &   .0099  &   .0066  &   	.1041  &  .01267  &   .0066  &   .34159  \\
.0000    &   	.0000  &   .0000  &   .2907  &   .1895  &   .0000  &   .0000  &   .1504  &   	.0000  &   .0000  &   .3624
\end{tabular} 
 \end{center} 

\par The problem of finding a solution is standard and there are a lot of different ways.  The easiest method is to interpolate all values for
 all columns, make negative numbers equal zero, find the sum of rows and normalize the rows by dividing by the sum.  If all variables are
independent, we can use the next prognosis formulas (2nd level, S $\rightarrow$ A):
 
\par x1=0;
\par x2=(-168.54121 + .08451year)/(-470.3754355 + .2358393year)
\par x3=(-218.045815 + .109309year)/(-470.3754355 + .2358393year)
\par x4=(-78.396752 + .0393176year)/(-470.3754355 + .2358393year)
\par x5=(-3.115175 + .0015597year)/(-470.3754355 + .2358393year)
\par x6=(-2.280141 + .001143year)/(-470.3754355 + .2358393year)

\par These formulas give us the ability of calculating the probability to use substitutions or deduction rules for the second level, case 
S $\rightarrow$  A$\_$ for an arbitrary time point. For instance, see below for a forecast of years 2000-02 (for year 2000 see row one, for
year 2001 see row two, for year 2002 see row three).
\vskip .2in
\par {\bf Prognosis  (2nd level, S $\rightarrow$  A, years 2000, 2001,2002,2003) }
\vskip .1in
  \begin {center} 
      \begin{tabular}{ ccccccc}
Year  &   AN*  &   DR  &   DR*  &     E$\_$  &   LP  &   S1  \\
2000  &	0.0000  &   	0.36741  &   	0.43907  &   	0.18298  &   	.0032421  &   .0044960  \\
2001  &   	0.0000  &   	0.36602  &   	0.44282  &   	0.18048  &   	.0037587  &   .0045497  \\
2002  &   	0.0000  &   	0.36500  &   	0.44556  &   	0.17865  &   	.0041381  &   .0045891  \\
2003  &   	0.0000  &   	0.36421  &   	0.44766  &   	0.17725  &   	.0044284  &   .0046193
\end{tabular}
 \end{center} 

\par We can compare real numbers and numbers calculated by our interpolation formulas (marked by sign ©):
 \begin {center}
    \begin{tabular} { ccccccc}
Year  &   AN*  &   DR  &   DR*  &   E$\_$  &   LP  &   S1  \\
1999    &   0.0000  &   0.3898  &   0.4550  &   0.1552  &   0.0000  &   0.0000  \\
1999  &    0.0000  &   0.36941  &   0.43368  &   0.18657  &   0.0024972  &   0.0044185
\end{tabular} 
\end{center} 
\par Note. The marginal probability is shown below (the set of probabilities when the year goes to infinity):
  \begin {center}
     \begin{tabular} { cccccc}
AN*  &   DR  &   DR*  &   E$\_$  &   LP  &   S1  \\
0.0000  &   	0.41939  &   	0.41939  &   	0.15085  &   	.0059842  &   .0043854
\end{tabular} 
 \end{center}   
\vskip .1in
\par {\bf 2nd level (S $\rightarrow$  DR) }
 \vskip .1in
\par We use similar methods as above to find the interpolation formulas for case S$\rightarrow$  DR, level 2:

\par x1=min(0,(-1.42842 + 0.000716year)/(3.78034 - .001393year))
\par x2=min(0,(-1.757595 + 0.000881year)/(3.78034 - .001393year))
\par x3=min(0,(-1.099245 + 0.000551year)/(3.78034 - .001393year))
\par x4=(81.861085 - 0.040811year)/(3.78034 - .001393year)
\par x5=min(0,(-78.91407 + 0.03959year)/(3.78034 - .001393year))
\par x6=( 6.80862 - 0.0034year)/(3.78034 - .001393year)
\par x7=(9.1221 - 0.00456year)/(3.78034 - .001393year)
\par x8=( 59.89132 - 0.02992year)/(3.78034 - .001393year)
\par x9=min(0,(-2.527665 + 0.001267year)/(3.78034 - .001393year))
\par x10=min(0,( -1.3167 + 0.00066year)/(3.78034 - .001393year))
\par x11=min(0,(-66.859091 + 0.03363year)/(3.78034 - .001393year))

\par The table below contains real figures and calculated by formulas (marked by symbol (c)).
\vskip .1in
  \begin {center}
    \begin{tabular} { lllllll}
Year  &   E$\_$  &   G$\_$  &   G$\_$S4  &   LP  &   LP*  &   RD  \\
1998(c)  &   .0021672  &   .0026667  &   .0016678  &   0.32358  &   0.18842  &   0.015558  \\
1998  &    	.00716  &   	.00881  &   	.00551  &   	0.30689  &   	0.1901  &    	0.0099  \\
1999(c)   &	.0028937  &   .0035605  &   .0022269  &   0.28280  &   0.22869  &   0.012145  \\
1999  &   .0000  &   .0000  &   .0000  &  .2907  &   .1895  &   .0000
\end{tabular} 
 \end{center}   
 \vskip .1in
 \begin {center}
    \begin{tabular} { llllll}
Year &   SO  &   S1  &   S1S4  &   	S2  &   S4  \\
1998(c)   &   	0.011320  &    0.11215  &   .0038350  &   .0019977  &   0.33663  \\
1998     &   	.0066  &   0.1041  &   .01267  &   .0066  &   0.34159  \\
1999(c)   &   	.0067291  &   0.082083 &   .0051206  &   .0026674  &   0.37109  \\
1999      &   	.0000  &   0.1504  &   .0000  &   .0000  &   0.3624
\end{tabular} 
 \end{center}   
\vskip .1in
\par The next table contains a prognosis for four years (2000 – 03).
\vskip .1in
 \begin {center}
    \begin{tabular} { lllllll}
Year	&  E$\_$  &   G$\_$  &   G$\_$S4  &   LP  &   LP*  &   RD   \\
2000  &   .0036222  &   .0044570  &   .0027875  &   0.24191  &   0.26907  &    .0087217  \\
2001  &   .0043420  &   .0053426  &   .0033414  &   0.20040  &   0.30879  &    .0052759  \\
2002  &   .0050066  &   .0061603  &   .0038528  &   0.15729  &   0.34474  &    .0018180   \\
2003  &   .0055303  &   .0068047  &   .0042558  &   0.11263  &   0.37142  &    .0000
\end{tabular} 
 \end{center}   

 \begin {center}
    \begin{tabular} { llllll}
Year &   SO  &   S1  &   S1S4  &     S2  &   S4   \\
2000 &   .0021248  &  0.051926  &   .0064097  &   .0033389  &  0.40564   \\
2001 &    .0000  &   0.021629  &   .0076834  &   .0040024  &   0.43919  \\
2002  &   .0000  &   0.0000  &   .0088594  &   	.0046150  &   0.46766   \\
2003  &   .0000  &   0.0000  &   .0097861  &   	.0050978  &   0.48448
\end{tabular} 
 \end{center}   
\par The last table contains (see below) the marginal probability:
 \begin {center}
    \begin{tabular} { ccccccccccc}
E$\_$  &   	G$\_$  &   G$\_$S4   &    LP  &    LP*  & 	RD  &   SO  &   S1  &   S1S4  &   S2  &   S4  \\
0.00  &   0.00  &   0.00  &   0.51862  &   0.00  &   0.043207  &   0.057948  &   0.38022  &   0.00  &   0.00  &   0.00
\end{tabular} 
 \end{center}   

\par The marginal probability provides us with a lot of information about tendency.  In this case, we see that only procedures,
coded by symbols LP, S1, SO, and RD, are essential for level 2, case S $ \rightarrow $  DR.

\par {\bf 1st level  }
\vskip .1in
\par We use the following formulas for a prognosis (see below).
\vskip .1in
\par x1=(-34.5413 + 0.017328year)/(-49.5236 + 0.025287year)
\par x2=(-38.1403 + 0.019139year)/(-49.5236 + 0.025287year)
\par x3=min(0,( 34.3195 - 0.016963year)/(-49.5236 + 0.025287year))
\par x4=(-81.3880 + 0.040860year)/(-49.5236 + 0.025287year)
\par x5=min(0, (12.2536 - 0.006126year)/(-49.5236 + 0.025287year))
\par x6=min(0,( 11.2749 - 0.005641year)/(-49.5236 + 0.025287year))
\par x7=( -0.7532 + 0.000378year)/(-49.5236 + 0.025287year)
\par x8=min(0, 26.5745 - 0.013270year)/(-49.5236 + 0.025287year))
\par x9=min(0,(  4.6698 - 0.002332year)/(-49.5236 + 0.025287year))
\par x10=(-0.1343 + 0.000075year)/(-49.5236 + 0.025287year)
\par x11=min(0,(15.2817 - 0.007632year)/(-49.5236 + 0.025287year))
\par x12=min(0,( 1.0595 - 0.000530year)/(-49.5236 + 0.025287year))

\par To check these formulas we compare numbers calculated by each formula with the given frequencies.  The table below
contains calculations (marked by symbol c) for the given year 1999.

 \begin {center}
    \begin{tabular} { ccccccc}
YEAR  &   A$\_$*  &   	A$\_$  &   DR  &   DR*  &   AN  &   AN*  \\
1999(c)  &   0.095037  &   	0.11572  &  	0.40062  &   	0.28416  &   	.0075407  &   0.0000  \\
1999  &   0.0589  &   0.1406  &   0.4470  &   0.2856  &   .00001  &   0.00001   \\
 \end{tabular} 
 \end{center}   

\vskip .2in
 \begin {center}
    \begin{tabular} { ccccccc}
YEAR  &   E$\_$	  &  LP  &   LP*  &   S1  &   S4  &   	RD   \\
1999(c)  &   .0023639  &   	0.046624  &   .0079370  &   0.015250  &   0.024724  &  0.000029280  \\
1999  &   .0000  &   0.0333  &   .0000  &   0.0208  &   0.0136   & 0.0000
 \end{tabular} 
 \end{center}    
\vskip .1in
\par { \bf The table below contains the numbers calculated by formulas for 2000-02.}
\vskip .1in
 \begin {center}
    \begin{tabular} {lllllll}
YEAR  &  A$\_$  &  A$\_$*  &  DR  &  DR*  &  AN  &  	AN*  \\
2000  &  0.10862  &  0.13040  &  0.37263  &  	0.31439  &   	.0015152  &  0.0000  \\
2001  &  0.12091  &  0.14363  &  0.34482  &  	0.34145  &   	0.0000  &   	0.0000  \\
2002  &  0.13222  &  0.15579  &  0.31833  &   	0.36626  &   	0.0000  &	0.0000  \\
2003  &  0.14140  &  0.16551  &  0.29063  &   	0.38561  &   	0.0000  &    	0.0000
\end{tabular} 
 \end{center}   
\vskip .1in
 \begin {center}
    \begin{tabular} { lllllll}
YEAR  &  E$\_$  &  LP  &  LP*  &  S1  &  S4  &  RD  \\
2000  &   .0026515  &   0.032670  &   .0054924  &   0.014867  &   	0.016761  &   0.0000  \\
2001  &   .0029103  &   0.019442  &   .0031759  &   0.014446  &   .0092199  &   0.0000  \\
2002  &   .0031481  &  .0070470    &   .0010057  &   0.014032  &	.0021566  &   0.0000  \\
2003  &   .0033371  &   0.0000      &   0.0000      &   0.013509  &   	0.0000      &   0.0000
\end{tabular} 
 \end{center}   
\vskip .1in
\par It is simple to find the marginal probabilities for the case when all variables are independent:
\vskip .1in
 \begin {center}
    \begin{tabular} { cccccccccc}
A$\_$   &           A$\_$*  &        DR  &      DR*  &      AN  &          AN*  &        E$\_$  &         LP  &           LP*  &         S1  \\
 0.22278  &     0.24607  &    0.0000  &     0.52533  &     0.0000  &     0.0000  &   .0048599  &    0.0000  &     0.0000  &    .00096426
\end{tabular} 
 \end{center}   
\vskip .1in

     \begin{tabular} { cc}
S4   &        RD    \\
0.0000  &    0.0000  
\end{tabular}  

\par This tells us that now there exist tendency towards marginal probability.  If this tendency remains safe
for a long time, doctors will only use the four procedures: DR,A$\_$, E$\_$, and S1.
 \vskip .1in
\par{\bf  An algorithm for the simulation of a list of short words }
 \vskip .1in
\par We can use the previous construction for the prognosis.  What is the significance of making a prognosis for non-numerical data? 
The significance is that we have to take the universal grammar and frequencies for the rules that are calculated by interpolation 
formulas, and then use model number I (forecast for the size of membership).  These details are what the real program consists of 
for the simulation.  We will obtain the number of steps from model I.
We first have to optimize our universal grammar.  After optimization, our universal grammar (one version is shown below) is divided 
into levels and combined with the calculations done before probabilities.

\par 1st level
 \vskip .1in 
\par {\bf  Table A }
 \vskip .1in
 \begin {center}
\begin{tabular} { |l|c|c|c|c|}
\hline
Substitution rules  &   2000  &   2001  &   2002  &   2003  \\
\hline
$ S  \rightarrow  A\_ * $  &   0.23902  &   0.26454  &   0.28801  &  	0.30691  \\
\hline
$ S  \rightarrow DR * $  &   0.68702  &   0.68627  &   0.68459  &   	0.67624  \\
\hline
$ S  \rightarrow AN * $  &   0.31439  &   0.0  &   0.0  &   0.0  \\
\hline
$ S  \rightarrow  E\_ $  &   	0.0026515  &   0.0029103  &   0.0031481  &   	0.0033371  \\
\hline 
$ S  \rightarrow LP  * $  &   0.0381624  &   0.022618  &   0.0080527  &   	0.0  \\
\hline
$ S  \rightarrow S1 $  &   	0.014867  &   0.014446  &   0.014032  &   0.013509  \\
\hline
$ S \rightarrow S4 $   &   	0.016761  &   0.0092199  &   0.0021566  &   	0.00  \\
\hline
$ S  \rightarrow RD $  &   	0.0  &   0.0  &   0.0  &   0.00  \\
\hline
\end{tabular}
\end{center}  

\par The symbol * means that there exists a substitution rule or a chain of rules that start from the last’s symbol given rule.  For instance,
 rule S  $\rightarrow  A\_$  has symbol * because there exists a rule or a sequence of rules which connect $A\_$ and another symbol (for instance AN) and so on.
The second table gives us the ability to stop (meaning that a one letter word was done and the algorithm goes to the next step of the
 loop) or start the generation of the next letter of word (this means that our word will contain at least two letters).

\par Table B
 \vskip .1in
 \begin {center}
\begin{tabular} { |l|c|c|c|c|}
\hline
After  &   2000  &   2001  &   2002  &   2003  \\
\hline
$ \rm A\_$ go to Stop  &   	0.45444  &   	0.457058  &   0.45908  &   0.46072  \\
\hline
$ \rm A\_$ go to 2nd leve  &   0.54556  &   0.542942  &  0.54092  &   0.53928  \\
\hline
DR go to Stop  &    	0.542386  &   0.50245  &   0.4650  &   0.42977  \\
\hline
DR go to 2nd level  &   0.457614  &   0.49755  &   0.53500  &   0.57023  \\
\hline
AN go to Stop  &   1  &   1  &   1  &   1  \\
\hline
AN go to 2nd level  &   0  &   0  &   0  &   0  \\
\hline
LP go to Stop  &   	0.85608  &   	0.85958  &   	0.87511  &   	1  \\
\hline
LP go to 2nd level  &   0.14392  &   0.14042  &   0.12489  &   0  \\
\hline
S1 go to Stop  &   	1  &   	1  &   	1  &   	1  \\
\hline
S1 go to 2nd level  &   0  &   0  &  0  &   0  \\
\hline
S4 go to Stop  &   	1  &   	1  &   	1  &   	1  \\
\hline
S4 go to 2nd level  &   0  &   0  &   0  &   	0  \\
\hline
\end{tabular}
\end {center}  

\par A reader starts with the left column.  For instance, the cell located at the first column and first row is “After $\rm A\_$ go to Stop” and so on. 
The probabilities are calculated by using the previous information. For example, the probability that event “after $ \rm A\_$ go to Stop” equals 
0.10862/0.23903=0.45444.  The opposite event “after $ \rm A\_$ go to 2nd level” equals (0.23903 - 0.10862)/0.23903 = 0.54556 or 1 – 0.45444 and so on.

\par {\bf  2nd level, case $ S \rightarrow  A\_$ }

\par Similarly, we calculate tables A2 and B2 for the second level (case $S \rightarrow  A\_$ ).
\vskip .1in
\par {\bf Table A2 }
 \vskip .1in
 \begin {center}
\begin{tabular} { |l|c|c|c|c|}
\hline
Substitution rules  &   2000  &   2001  &   2002  &   2003  \\
\hline
$ A\_ \rightarrow AN $  &   0  &   0  &   0  &   0  \\
\hline
$ A\_\rightarrow DR* $  &   0.80648  &   0.80884  &   0.81056  &   0.81176  \\
\hline
$ A\_ \rightarrow E\_ $  &  	0.18298  &   	0.18048  &   	0.17865  &   	0.177725  \\
\hline
$ A\_ \rightarrow LP $  &   0.0032421  &   0.0037587  &   0.0041381  &   	0.0044284  \\
\hline
$ A\_ \rightarrow S1 $  &  	0.004496  &   0.0045497  &   0.0045891  &   0.0046193  \\
\hline
\end{tabular}
\end{center}  

\vskip .1in
\par {\bf Table B2}
 \vskip .1in
 \begin {center}
\begin{tabular} { |l|c|c|c|c|}
\hline 
After  &   2000  &   2001  &   2002  &   2003  \\
\hline
DR go to Stop  &   0.455557  &   0.45324  &   0.4503  &   0.448667  \\
\hline
DR go to 3rd level  &   0.54443  &   0.54676  &   05497  &   0.55133  \\
\hline
\end{tabular}
\end{center}  

\par A lot of levels exist in the real model and for these levels it is necessary to prepare tables A and B.  (We do not calculate tables 
A and B any more because it is not important for our understanding.)   So, suppose we have the hierarchy system of tables A and B. 
 The number of pairs in the tables must equal the number of levels. The algorithm is a loop and every step has to start from the 
first level.  After “Stop,” the algorithm prints a word and goes back to first level.  On level N a special program randomly gets a
 substitution rule from the first column of Table A (with probability from the right part of the Table A).  Then, if the rule does not
 have the special symbol *, the algorithm prints out a word and goes to the first level.  If symbol * is present, the algorithm prints 
an additional letter (right part of rule if right part is a term) and goes to table B.  The algorithm then randomly goes to Stop or to Table A
 of level N +1, and so on.
\par Thereupon, we can finish the description of the process making of the semantic model of internal flow. We hope that the given algorithm may be used in the modeling of other companies.

\par {\bf CONCLUSION}
\par This paper represents the philosophy or point of view of a company as a device that has used the set of languages based 
on parts, natural processes, human actions, pose, etc. as an alphabet.  In this article we emphasize on informational processes, 
and firstly on the communication process.  Similarly, we pay more attention to the classical point of view on human beings as a
 communication system and try to describe all languages used by humans.
\par All companies are broken into two parts in this article: industrial and purely informational companies. The industrial company 
(industrial part of a company’s activities) is modeled as an input flow of words where any word represents assembling or 
disassembling actions.  Any word enables us to generate the company’s structure, figure out the number of employees, the 
movement’s path, and so on.  The creation of a semantic space for industries is beyond our abilities and depends on a particular 
industry.  This article shows us only a hint of how to do this.  We show that the more deeply developed languages of parts,
processes, actions plus description of shapes and conditions, and technological semantic space give us the ability to build more 
realistic models of industrial companies by using the described method.
\par A pure informational company is represented as a model of input flow of an insurance company.  In real life all companies 
have industrial and informational parts.  Companies in which the industrial part is the main section are called “industrial” companies.  
Companies where informational business is the main topic are called “informational” companies.  But for both types of companies, 
and in every day life, conversations between people are important functions.
\par The language philosophy is used for the modeling of communication as a random walk in the semantic space.  In this article 
a representation of an approach to build the semantic space as tree is demonstrated, where concepts are nodes and more 
general concepts have higher positions in the tree than less general concepts.  We then build the semantic tree and write 
the system of differential equations.  The system of differential equations describes an evolution of conversations. 
 The description of conversations depends on the type of conversation (business, sport, life, and so on), the shape of 
the table, and other factors.  A few cases are analyzed: conversations with a square table and free conversations.  
The represented method may also be used in the modeling of conflicts.
\par We then do an example of modeling certain features of a pure informational company: an insurance company.  One of 
main purposes of this modeling is to be able to find the semantic (linguistic) base of an insurance company.  The model of an 
insurance company is done as a computer simulation of the input flow.  The keystone of this simulation is a semantic model 
of the history of clients' diseases.  To construct this semantic model we use the alphabet of the types of procedures performed 
on patients with the given chronic disease.  For instance, the alphabet for diabetes contains 34 "letters”.  The history of disease 
can be represented by a short word in a given alphabet and looks like "A$\_$ANDR S4 S2", where A$\_$, AN,$\ldots$ , S4, S2 are letters 
of the alphabet of the disease.  The study of the information for five years shows us that the structure of short words has a 
tendency to change.  To model this tendency we use Markov chains. The conditional probability is found from the data. Then, 
using a computer simulation, we calculate a set of pseudo-random "short words.”  The next problem shown is the generation of 
the set of pseudo-random "long words.”  The long word may look like "A$\_$-12AN-1DR-17 S4-1S2-1", where the number that 
follows each "letter" is the frequency of encountering this letter. For this purpose we find the conditional probability 
Pr{X=" long word"/X=" short word"} and then generate a set of pseudo-random "long words.” The list of "long words" and 
the list of "normative prices" for procedures give us the ability to calculate the mean, "harmonic", minimal and maximal prices 
for all diseases. This model may be used when making the forecast of an insurance company. This model is the basis for a
more comprehensive model of an insurance company. 
\par To predict the number of patients we construct a special system of differential equations and then from these equations 
we derive a statistical model suitable for computer simulation.  The author believes that the representation given in this paper 
(the point of view on an insurance company as a device that uses the set of natural processes where this process communicates) 
can be spread on all types of companies for all industries.  It is the main reason why the description of internal flow for an 
insurance company is presented with big details. 
\par In the future the author plans to make a universal model of the external world for a company and work out the whole 
semantic model of a company.  Section three’s model can be used in the decision - making process.  People responsible for 
some type of disease or group of diseases must keep in mind the meaning of a “normal” word (normal sequence or sequences 
of procedures) for a given diagnosis. If the word does not belong to the normal and/or more often occurrence set of words, 
the employee begins an investigation (not criminal). For this function (recognition of situation) we use the special self-learning 
grammar. For numerical variables we can use a neural networks algorithm, but for our case we use grammar.

\bibliographystyle{amsplain}

\par {\bf Appendix A}
\par This appendix contains training corps (Table 1 –3, 5) for getting formal grammar for “procedure language”. All tables contain a list of 
words with probabilities (frequencies) greater than 0.02 only. In reality all words are used without restrictions. Table 4 contains words
with probabilities greater than 0.001 and we essentially see longer words. For every training corps we deduce the formal grammar and 
then find the tendency of frequencies.    
 \vskip .2in
\par { \bf Table 1} (1995:812 patients)
 \vskip .1in
 \begin {center}
 \begin{tabular} { |l|c|c|l|}
\hline
WORD &  NUMBER PATIENTS & PROBABILITY & EXPLANATION \\
\hline
A$\_$   &   19   &    0.0234   &	TRANSPORT     \\
\hline
A$\_$AN  &   24   &   0.0295   &  T+ANESTHESIA  \\
\hline
AN   &  94  &   0.11576 &	ANESTHESIA   \\
\hline
ANDR	& 17 &  0.0209 &	ANEST+DRUG \\
\hline
DR  &	357 &   0.4396  &	DRUG  \\
\hline
DRLP & 35 &	0.0431  &	DR+LAB PATOLOGY  \\
\hline
DRS1 & 25 &  0.03078 &	DR+SURG:INTEG \\
\hline
DRS4 & 17 &  0.0209 &	DR+SURG:CARD  \\
\hline
LP  &	51 &	0.0628 &	LAB  PATOLOGY \\
\hline 
S1  &	18  &	0.02216  &	SURG:INTEG \\
\hline
S4  &	19  &	0.0234 &	SURG:CARD \\
\hline
\end{tabular}
\end{center} 
		
\par The whole population contains 812 patients. This table contains 676 patients or 83.25$\%$ of whole population.
 \vskip .2in
 \par { \bf Table 2 } (1996:2097 patients)
  \vskip .1in
 \begin {center}
 \begin{tabular} { |l|c|c|l|}
\hline
WORD &  NUMBER PATIENTS & PROBABILITY & EXPLANATION \\
\hline
A$\_$   &	60  &	0.0286 &	TRANSPORT \\
\hline
A$\_$AN   &	40  &	0.01907  &	T+ANESTHESIA \\
\hline
A$\_$ANDR  &	29  &	0.01383  &	T+ANESTH+DRUG  \\
\hline
A$\_$DR  &	26  &	0.0124  &	T+DRUG  \\
\hline
A$\_$DRE$\_$  &	14  &	0.00667  &	T+DRUG+ DIGEST  \\
\hline
A$\_$DRLP  &	12  &	0.00572  &	T+DRUG+LAB PATO  \\
\hline
A$\_$E$\_$   &     13  &	0.0062  &	T+ DIGEST  \\
\hline
AN  &	83  &	0.03958  &	ANESTHESIA  \\
\hline
ANDR &   25 &  0.01192 &	ANESTHES+DRUG  \\
\hline
DR &	828  &  0.3948  &	DRUG  \\
\hline
DRLP &  141 &  0.0672  &	DRUG+LAB PATOL  \\
\hline
DRLPS4  &	63  &	0.0300  &	DR+LB+SURG:CARD \\
\hline
DRRD  &  16	&  0.00762  &  DRUG+RAD:DIAG  \\
\hline
DRSO  &  19  &  0.00906  &  DRUG+SURG:EYE \\
\hline
DRS1  &  20  &  0.00954  &  DRUG+SURG:INTEG  \\
\hline
DRS4  &  105  &  0.05007  &  DRUG+SURG:CARD  \\
\hline
LP  &	187 &  0.08917  &  	LAB  PATOLOGY \\
\hline
LPS4  &  55  &  0.02622  &  LAB+SURG:CARD  \\
\hline
RD  &	10  &	0.00477  &	RAD:DIAG  \\
\hline
S1  &	15  &	0.00954  &	SURGERY:INTEG  \\
\hline
S4  &	126  &  0.06008  &	SURGERY:CARD \\
\hline
\end  {tabular} 
\end {center}	
 \vskip .2in
\par {\bf Table 3 } (1997: 3067 patients)
 \vskip .1in
\begin {center}
 \begin{tabular} { |l|c|c|l|}
\hline
WORD &  NUMBER PATIENTS & PROBABILITY & EXPLANATION \\
\hline
A$\_$   &	143  &  0.0466  &  TRANSPORT  \\
\hline
A$\_$AN  &  65  &  0.02119  &  TRAN+ ANESTHESIA  \\
\hline
A$\_$ANDR  &	35  &	0.0114  &	T+ANESTH+DRUG  \\
\hline
A$\_$DR  &  80  &  0.02608  &  T+DRUG  \\
\hline
A$\_$DRE$\_$   &	27  &	0.00880  &	T+DRUG+DIGEST  \\
\hline
A$\_$DRL$\_$   &	14  &	0.00456  &	T+DRUG+LAB PAT  \\
\hline
A$\_$DRS4   &	14  &	0.00456  &	T+DRG+SURG:CARD  \\
\hline
A$\_$E$\_$   &  48  &  0.01565  &  T+DIGEST  \\
\hline
AN  &  95  &  0.03097  &  ANESTHESIA  \\
\hline
ANDR  &  15  &  0.00489  &  ANESTHES+DRUG  \\
\hline
DR  &	1047  &  0.34137  &  DRUG  \\
\hline
DRLP  &  	201  &  0.0655  &  DRUG+LAB PAT  \\
\hline
DRLPS4  &  	100  &  0.0326  &  DRG+LB+SRG:CARD  \\
\hline
DRRD  &  22  &  0.0072  &  DRUG+RAD:DIAG  \\
\hline
DRSO  &  11  &  0.00358  &  DRUG+SURG:EYE \\
\hline
DRS1  &  39  &  0.0127  &  DRUG+SURG:INTEG  \\
\hline
DRS4  &  173  &  0.0564  &  DRUG+SURG:CARD  \\
\hline
E$\_$  &  	13  &  	0.00424  &  DIGEST  \\
\hline
LP  &  	314  &  0.10238  &   LAB PATOLOGY \\
\hline
LPS4  &  80  &  0.02608  &  LAB+SURG:CARD  \\
\hline
M$\_$  &  15  &  0.00489  &  MED  SERVICE  \\
\hline
S1  &  17  &  	0.00554  &  SURG:INTEG  \\
\hline
S4  &  	208  &  0.06782  &  SURG:CARD  \\
\hline
\end{tabular} 
\end{center}   
This table contains 2776 patients what is 90.51$\%$ of whole population (3067 patients). 
 \vskip .2in
\par {\bf Table 4 }(1998: 7452 patients)
\vskip .1in
  \begin {center}
 \begin{tabular} { |l|c|c|l|}
\hline
WORD &  NUMBER PATIENTS & PROBABILITY & EXPLANATION \\
\hline
A$\_$   &  367  &  0.04925  &  TRANSPORT  \\
\hline
A$\_$AN  &  61  &  0.008185  &  T+ ANESTHESIA  \\
\hline
A$\_$ANDR  &  	57  &  	0.007648  &  T+ANEST+DRUG  \\
\hline
A$\_$ANDRE$\_$   &  23  &  0.003086  &  T+AN+DRG+DIGEST  \\
\hline
A$\_$ANDRLP  &  17  &  0.00228  &  T+AN+DR+LAB PAT  \\
\hline
A$\_$ANDRS4  &  15  &  0.00201  &  T+A+D+SURG:CARD \\
\hline
A$\_$DR  &  255  &  0.0342  &  T+DRUG  \\
\hline
A$\_$DRE$\_$  &  	139  &  0.01865  &  T+DRUG+DIGEST  \\
\hline
A$\_$DRE$\_$LP  &  19  &  0.00255  &  	T+DR+DIGEST+LAB  \\
\hline
A$\_$DRE$\_$LPS4  &  18  &  0.00241  &  +SURGERY:CARD  \\
\hline  
A$\_$DRE$\_$S1  &  10  &  0.00134  &  T+DR+DIG+SRG:INT \\
\hline
A$\_$DRE$\_$S4  &  28  &  0.00375  &  +SURGERY:CARD  \\
\hline
A$\_$DRLP  &  	48  &  	0.00644  &  T+DRG+LAB PATOL  \\
\hline
A$\_$DRLPS4  &  19  &  0.00254  &  	+ SURGERY:CARD  \\
\hline
A$\_$DRS1  &  	28  &  	0.00375  &  	T+DRG+SURG:INTEG  \\ 
\hline
A$\_$DRS4  &  	62  &  	0.00831  &  	T+DRG+SURG:CARD  \\
\hline
A$\_$E$\_$   &  182  &  0.0244  &  T+DIGEST  \\
\hline
A$\_$LP  &  16  &  0.00214  &  T+LAB PATOLOGY  \\
\hline
A$\_$S1  &  12  &  0.00161  &  T+SURGERY:INTEG  \\
\hline
AN  &  74  &  0.00993  &  ANESTHESIA  \\
\hline
ANDR  &  40  &  0.00536  &  ANESTHESIA+DRUG  \\
\hline
ANDRS4  &  	11  &  	0.00147  &  +SURGERY:CARD  \\
\hline
DR  &  3032  &  0.40687  &  DRUG  \\
\hline
DRE$\_$   &  13  &  0.00174  &  DRUG+DIGEST  \\
\hline
DRG$\_$  &  16  &  0.00214  &  DRUG+PRF SERVICE  \\
\hline
DRG$\_$S4  &  10  &  0.00134  &  +SURGERY:CARD  \\
\hline
DRLP  &  557  &  0.074745  &  DRUG+LAB PAT  \\
\hline
DRLPRD  &  12  &  0.00161  &  +RAD:DIAG  \\
\hline
DRLPRDS4  &  10  &  0.00134  &  +SURGERY:CARD  \\
\hline
DRLPS1  &  31  &  0.00416  &  DR+LAB+SRG:INTEG \\
\hline
DRLPS1S4  &  11  &  0.00146  &  	+SURGERY:CARD  \\
\hline
DRLPS4  &  	281  &  0.0377  &  	DR+LAB+SRG:CARD  \\
\hline
DRRD  &  18  &  0.002415  &  DR+RAD:DIAG  \\
\hline
DRSO  &  12  &  0.00161  &  DRG+SURG:EYE  \\
\hline
DRS1  &  189  &  0.02536  &  DRUG+SURG:INTEG  \\
\hline
DRS1S4  &  23  &  0.00309  &  +SURG:CARD  \\
\hline
DRS2  &  12  &  0.00161  &  DRUG+SRG:MUSC  \\
\hline
DRS4  &  620  &  0.0832  &  DRUG+SRG:CARD  \\
\hline
E$\_$  &  	27  &  	0.00362  &  DIGEST  \\
\hline
LP  &  	356  &  0.04777  &  LAB  PATOLOGY  \\
\hline
LPS4  &  42  &  0.005636  &  LAB+SURG:CARD  \\
\hline
RD  &  10  &  0.00134  &  	RAD:DIAG  \\
\hline
S1  &  	122  &  0.01637  &  SURGERY:INTEG  \\
\hline
S4  &  	137  &  0.01838  &  SURGERY:CARD  \\
\hline
\end{tabular}
\end{center}	
\par This table contains 7032 patients what is 94.36$\%$ of whole population (7452 patients).
 \vskip .2in
\par {\bf Table 5 }  ( 1999: 11730 patients)   
 \vskip .1in
 \begin {center}
\begin{tabular} { |l|c|c|l|}
\hline
WORD &  NUMBER PATIENTS & PROBABILITY & EXPLANATION \\
\hline
A$\_$   &  591   &  0.05038   &  TRANSPORT  \\
\hline
A$\_$DR   &  550   &  0.04689   & 	TRAN+DRUG \\
\hline
A$\_$DRE$\_$    &  297   &   0.02532   &  DIGEST  \\
\hline
A$\_$DRLP   & 	123   &  0.01083   &   LAB PAT \\
\hline
A$\_$DRS1   & 	95   &  0.0081   &  	SURG:INTEG  \\
\hline
A$\_$DRS4   & 	127   &  0.010827   &  SURG:CARD  \\
\hline
A$\_$E$\_$   &  219   &   0.01867   &  	DIGEST  \\
\hline
DR   &   4486   &   0.38244   &   	DRUG  \\
\hline
DRLP   &   839   &   	0.071526   &    LAB  PATOLOGY  \\
\hline
DRLPS4   &   	547   &  0.0466   &   SURG:CARD  \\
\hline
DRS1   &   434   &   	0.037    &   SURG:INTEG  \\
\hline
DRS4   &    1046   &    0.089   &	SURG:CARD  \\
\hline
LP   &  334   &  0.02847   &  LAB  PATOLOGY  \\
\hline
S1   &   209   &  0.01782   &   SURG:INTEG  \\
\hline
S4   & 	 137   &   0.01168   &   SURG:CARD  \\	
\hline
\end{tabular}
\end{center}

\end{document}